\documentclass{article}

\usepackage{amsmath,amsfonts,bm}

\def\Secref#1{Section~\ref{#1}}

\def\eqref#1{equation~\ref{#1}}

\def\1{\bm{1}}

\def\vtheta{{\bm{\theta}}}

\def\ve{{\bm{e}}}

\def\vu{{\bm{u}}}
\def\vv{{\bm{v}}}

\def\vx{{\bm{x}}}

\def\mE{{\bm{E}}}

\def\mI{{\bm{I}}}

\def\mO{{\bm{O}}}
\def\mP{{\bm{P}}}

\def\mU{{\bm{U}}}
\def\mV{{\bm{V}}}
\def\mW{{\bm{W}}}

\def\mSigma{{\bm{\Sigma}}}

\DeclareMathAlphabet{\mathsfit}{\encodingdefault}{\sfdefault}{m}{sl}
\SetMathAlphabet{\mathsfit}{bold}{\encodingdefault}{\sfdefault}{bx}{n}

\usepackage{microtype}
\usepackage{graphicx}
\usepackage{subfig}
\usepackage{booktabs} 

\usepackage{url}
\usepackage{hyperref}
\usepackage{multirow}
\usepackage{wrapfig}
\usepackage{tcolorbox}

\usepackage{hyperref}

\usepackage[accepted]{icml2024}

\usepackage{amsmath}
\usepackage{amssymb}
\usepackage{mathtools}
\usepackage{amsthm}

\usepackage[capitalize,noabbrev]{cleveref}

\theoremstyle{plain}
\newtheorem{theorem}{Theorem}[section]

\theoremstyle{definition}
\newtheorem{definition}[theorem]{Definition}

\theoremstyle{remark}

\newcommand{\lms}[1]{{#1}}

\usepackage[textsize=tiny]{todonotes}

\icmltitlerunning{On the Embedding Collapse When Scaling Up Recommendation Models}

\begin{document}

\twocolumn[
\icmltitle{On the Embedding Collapse When Scaling Up Recommendation Models}




\begin{icmlauthorlist}
\icmlsetsymbol{thanks}{*}
\icmlauthor{Xingzhuo Guo}{sch,thanks}
\icmlauthor{Junwei Pan}{comp}
\icmlauthor{Ximei Wang}{comp}
\icmlauthor{Baixu Chen}{sch}
\icmlauthor{Jie Jiang}{comp}
\icmlauthor{Mingsheng Long}{sch}
\end{icmlauthorlist}

\icmlaffiliation{sch}{School of Software, BNRist, Tsinghua University, China}
\icmlaffiliation{comp}{Tencent Inc, China. Xingzhuo Guo \textless{}gxz23@mails.tsinghua.edu.cn\textgreater{}}

\icmlcorrespondingauthor{Mingsheng Long}{mingsheng@tsinghua.edu.cn}

\icmlkeywords{Machine Learning, ICML}

\vskip 0.3in
]



\printAffiliationsAndNotice{\textsuperscript{*}Work partially done while an intern at Tencent Inc.}  

\begin{abstract}
    Recent advances in foundation models have led to a promising trend of developing large recommendation models to leverage vast amounts of available data. \lms{Still}, mainstream models \lms{remain embarrassingly small in} size and na\"ive enlarging does not lead to sufficient performance gain, suggesting a deficiency in the \emph{model scalability}. In this paper, we identify the \emph{embedding collapse} phenomenon as the inhibition of scalability, wherein the embedding matrix tends to occupy a low-dimensional subspace. Through empirical and theoretical analysis, we demonstrate a \emph{two-sided effect} of feature interaction specific to recommendation models. On the one hand, interacting with collapsed embeddings restricts embedding learning and exacerbates the collapse issue. On the other hand, interaction is crucial in mitigating the fitting of spurious features as a scalability guarantee. Based on our analysis, we propose a simple yet effective \emph{multi-embedding} design incorporating embedding-set-specific interaction modules to learn embedding sets with large diversity and thus reduce collapse. Extensive experiments demonstrate that this proposed design provides consistent scalability and effective collapse mitigation for various recommendation models. Code is available at this repository: \url{https://github.com/thuml/Multi-Embedding}.
\end{abstract}


\section{Introduction}

Recommender systems are \lms{important} machine learning scenarios that predict users' actions on items based on tremendous multi-field categorical data~\citep{mutli_field_cat}, which play an indispensable role in our daily lives to help people discover information about their interests and have been adopted in a wide range of online applications, such as E-commerce, social media, news feeds, and music streaming. Researchers have developed deep-learning-based recommendation models to dig feature representations flexibly. These models have been successfully deployed across a multitude of application scenarios, thereby demonstrating their widespread adoption and effectiveness.

\begin{figure*}[ht]
    \centering
    \begin{minipage}{0.313\textwidth}
        \centering
        \subfloat[Performance when scaling up recommendation models\label{subfig:intro-a}]{%
        \includegraphics[width=0.95\textwidth]{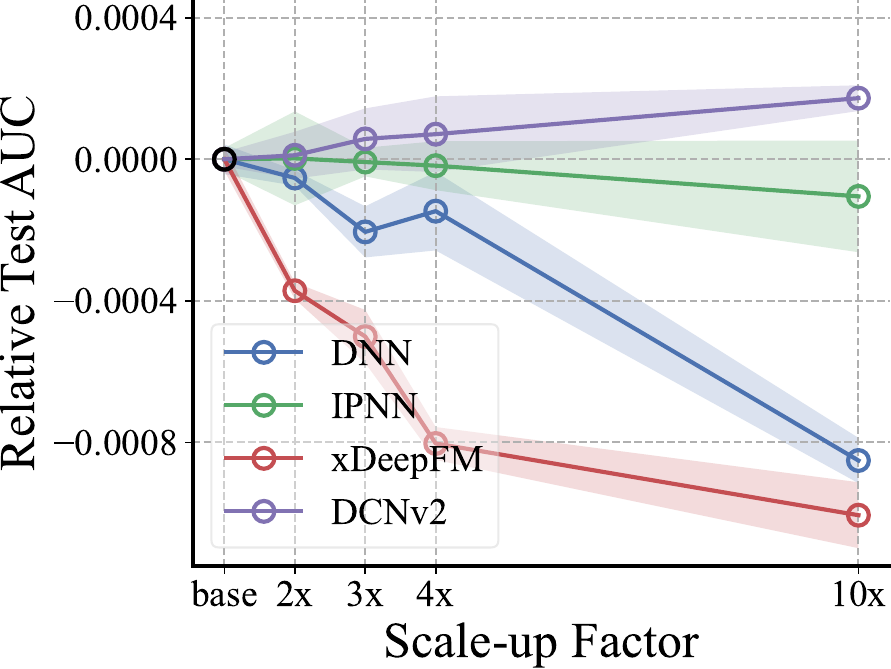}%
        }
    \end{minipage}%
    \begin{minipage}{0.687\textwidth}
        \centering
        \subfloat[Singular values of DCNv2 under different model size\label{subfig:intro-b}, with the dashed lines corresponding to the base size.]{
            \begin{tabular}{ccc}
                \includegraphics[width=0.27\textwidth]{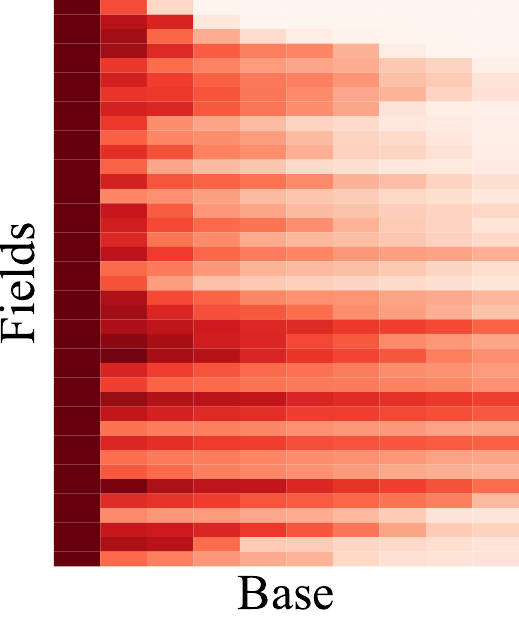}&
                \includegraphics[width=0.27\textwidth]{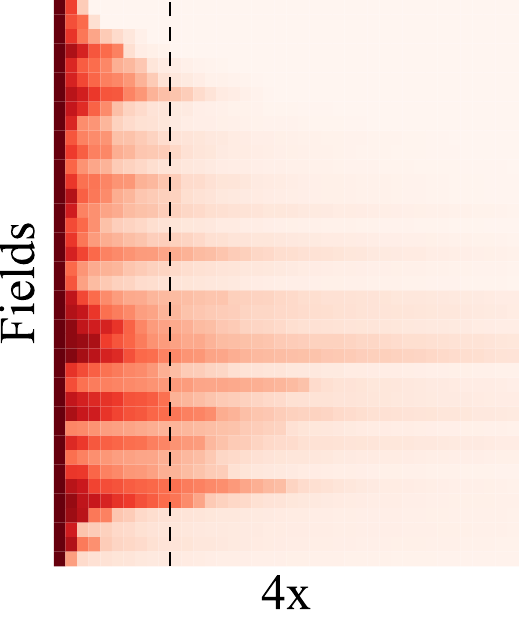}&
                \includegraphics[width=0.27\textwidth]{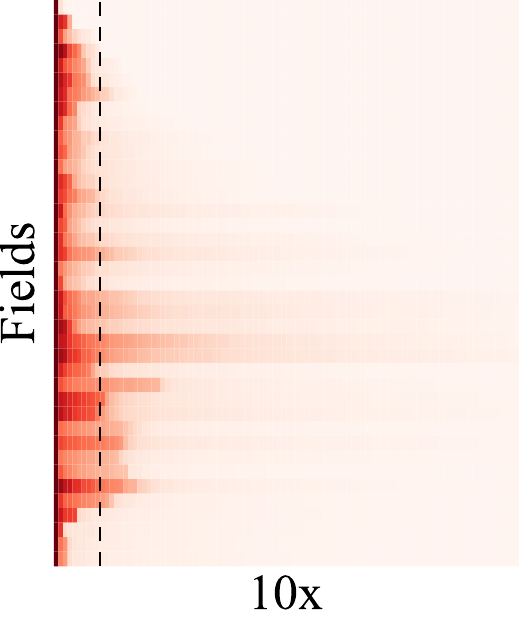}%
            \end{tabular}
        }
    \end{minipage}%
    \caption{Unsatisfactory scalability of existing recommendation models. \textbf{(a)}: Increasing the embedding size does not improve remarkably or even hurts the model performance. \textbf{(b)}: Most embedding matrices do not learn large singular values and tend to be low-rank.}
    \label{fig:intro}
\end{figure*}

Motivated by the advancement of large foundation models~\citep{sam,gpt4,clip,stablediffusion} that benefit from increasing parameters, it should have been a promising trend to scale up the recommendation model size to use the data amount fully. Yet counterintuitively, as the most-weighted component critical for performance~\citep{ipnn,xdeepfm,dcnv2}, the embeddings of recommendation models are typically tuned too small such as a size of 10~\citep{bars}, and thus do not adequately capture the magnitude of data. Worsely, increasing the embedding size does not sufficiently improve the performance or even hurts the model, as shown in Figure~\ref{subfig:intro-a}. This suggests a deficiency in the \emph{model scalability} of existing architecture designs, constraining the potential upper bound for recommender systems.

To figure out the reason behind it, we take a spectral analysis on the learned embedding matrices based on singular value decomposition and exhibit the normalized singular values in Figure~\ref{subfig:intro-b}. Surprisingly, most singular values are significantly small, \textit{i.e.}, the learned embedding matrices are nearly low-rank, which we refer to as the \emph{embedding collapse} phenomenon. With the enlarged model size, the model does not learn to capture a larger dimension of information, implying a learning process with ineffective parameter utilization, which restricts the scalability.

In this work, we study the mechanism behind the embedding collapse through empirical and theoretical analysis and shed light on the \emph{two-sided effect} on model scalability of the feature interaction module, the \lms{cornerstone} of recommendation models to model higher-order correlations. On the one hand, interaction with collapsed embeddings will constrain the embedding learning and, thus, in turn, aggravate the collapse issue. On the other hand, the feature interaction also plays a vital role in reducing overfitting when scaling up models, which cannot be restricted or removed.

Based on our analysis, we conclude the principle to mitigate collapse without suppressing feature interaction, \lms{so that} scalable models \lms{can be approached}. We propose \emph{multi-embedding} as a simple yet efficient design for model scaling. Multi-embedding scales the number of independent embedding sets and incorporates embedding-set-specific interaction modules to jointly capture different patterns. Our experimental results demonstrate that multi-embedding provides scalability for extensive mainstream models and significantly mitigates embedding collapse, pointing to a methodology of breaking through the \emph{size limit} of recommender systems.

Our contributions can be summarized as:
\begin{itemize}
    \item To the best of our knowledge, we are the first to point out the \emph{model scalability} issue in recommender systems and discover the \emph{embedding collapse} phenomenon, an urgent problem to address to enhance scalability.
    \item Using empirical and theoretical analysis, we shed light on the \emph{two-sided effect} of the feature interaction process on scalability based on the collapse phenomenon. We reveal that feature interaction leads to collapse while providing essential overfitting \lms{resistance}.
    \item Following our concluded principle to mitigate collapse without suppressing feature interaction, we propose \emph{multi-embedding} as a simple unified design, consistently improving scalability and effectively \lms{mitigating} embedding collapse for extensive state-of-the-art recommendation models.
\end{itemize}

\section{Preliminaries}

Recommendation models aim to predict an action based on features from various fields. Throughout this paper, we consider the fundamental scenario of recommender systems, in which categorial features and binary outputs are involved. Formally, suppose there are $N$ fields, with the $i$-th field denoted as $\mathcal{X}_i=\{1,2,...,D_i\}$, where $D_i$ denotes the field cardinality. Let
\[
\mathcal{X}=\mathcal{X}_1\times\mathcal{X}_2\times...\times\mathcal{X}_N
\]
and $\mathcal{Y}=\{0,1\}$, then recommendation models aim to learn a mapping from $\mathcal{X}$ to $\mathcal{Y}$. In addition to considering individual features from diverse fields, there have been numerous studies~\citep{mf, fm, ffm, deepfm, xdeepfm, fwfm, fmfm, dcnv2} within the area of recommender systems to model combined features using \emph{feature interaction} modules. In this work, we investigate the following widely adopted architecture for mainstream models. A model comprises: (1) embedding layers $\mE_i\in\mathbb{R}^{D_i\times K}$ for each field, with embedding size $K$; (2) an interaction module $I$ responsible for integrating all embeddings into a combined feature scalar or vector; and (3) a subsequent postprocessing module $F$ used for prediction purposes, such as MLP and MoE. The forward pass of such a model is formalized as
\begin{align*}
    \ve_i&=\mE_i^{\top}\1_{x_i},\ \forall i\in\{1,2,...,N\}, \\
    h&=I(\ve_1,\ve_2,...,\ve_n), \\
    \hat{y}&=F(h),
\end{align*}
where $\1_{x_i}$ indicates the one-hot encoding of $x_i\in\mathcal{X}_i$, in other words, $\ve_i$ refers to (transposed) $x_i$-th row of the embedding table $\mE_i$.

\section{Embedding Collapse}

Singular value decomposition has been widely used to measure the collapse phenomenon~\citep{jing2021understanding}. In Figure~\ref{subfig:intro-b}, we have shown that the learned embedding matrices of recommendation models are approximately low-rank with some extremely small singular values. To determine the degree of collapse for such matrices with low-rank tendencies, we propose \emph{information abundance} as a generalized quantification. 

\begin{definition}[Information Abundance]
    Consider a matrix $\mE\in\mathbb{R}^{D\times K}$ and its singular value decomposition $\mE=\mU\mSigma\mV=\sum\limits_{k=1}^K\sigma_k\vu_k\vv_k^\top$, then the \emph{information abundance} of $\mE$ is defined as

    \[
    \mathrm{IA}(\mE)=\frac{\|\bm{\sigma}\|_1}{\|\bm{\sigma}\|_\infty},
    \]
    \textit{i.e.}, the sum of all singular values normalized by the maximum singular value.
\end{definition}

\begin{figure}[ht]
    \centering
    \includegraphics[width=0.5\linewidth]{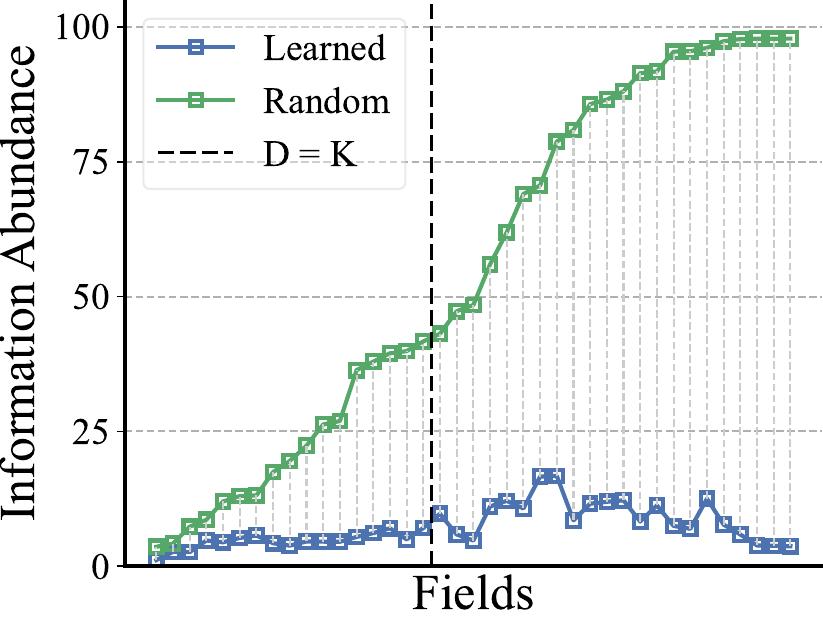}
    \caption{Visualization of information abundance on the Criteo dataset. Fields are sorted by their cardinalities.}
    \label{fig:ia-vis}
\end{figure}

Intuitively, a matrix with high information abundance demonstrates a balanced distribution in vector space since it has similar singular values. In contrast, a matrix with low information abundance suggests that the components corresponding to smaller singular values can be compressed without significantly impacting the result. Compared with matrix rank, information abundance can be regarded as a simple extension by noticing that $\mathrm{rank}(\mE)=\|\bm{\sigma}\|_0$, yet it is applicable for non-strictly low-rank matrices, especially for fields with $D_i\gg K$ which is possibly of rank $K$. We calculate the information abundance of embedding matrices for the enlarged DCNv2~\citep{dcnv2} and compare it with that of randomly initialized matrices, shown in Figure~\ref{fig:ia-vis}. It is observed that the information abundance of learned embedding matrices is extremely low, indicating the embedding collapse phenomenon.

\section{Feature Interaction Revisited}
\label{sec:fi}

In this section, we delve deeper into the embedding collapse phenomenon for recommendation models. Our investigation revisits feature interaction modules which are the key to recommendation models, and revolves around two questions: (1) How is embedding collapse caused? (2) How to properly mitigate embedding collapse \lms{and} enhance scalability? Through empirical and theoretical studies, we shed light on the two-sided effect of feature interaction modules on model scalability.

\subsection{Interaction-Collapse Theory}

To determine how feature interaction leads to embedding collapse, it is inadequate to directly analyze the raw embedding matrices since the learned embedding matrix results from interactions with all other fields, making it difficult to isolate the impact of field-pair-level interaction on embedding learning. Under this obstacle, we propose empirical evidence on models with \emph{sub-embeddings} and theoretical analysis on general models, and conclude that feature interaction causes embedding collapse, named the \emph{interaction-collapse theory}.

\begin{figure*}[t]
    \centering
    \begin{minipage}{0.72\textwidth}
        \begin{minipage}{0.95\textwidth}
            \centering
            \begin{minipage}{0.3333\textwidth}
                \centering
                \subfloat[$\mathrm{IA}(\mE_i^{\to j})$.\label{subfig:evd2d}]{\includegraphics[width=0.95\linewidth]{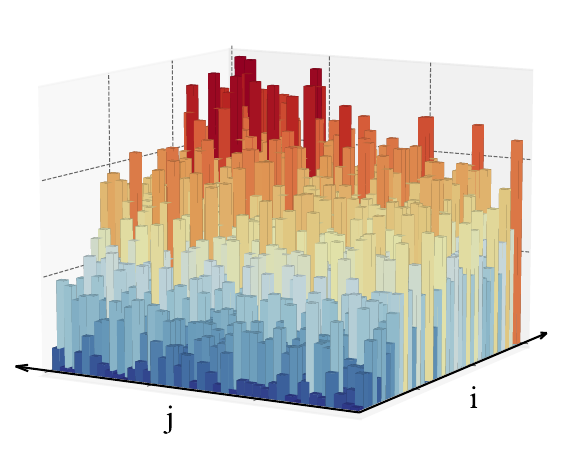}}%
            \end{minipage}%
            \begin{minipage}{0.3333\textwidth}
                \centering
                \subfloat[$\sum_{j=1}^N\mathrm{IA}(\mE_i^{\to j})$.\label{subfig:evd2e}]{\includegraphics[width=0.95\linewidth]{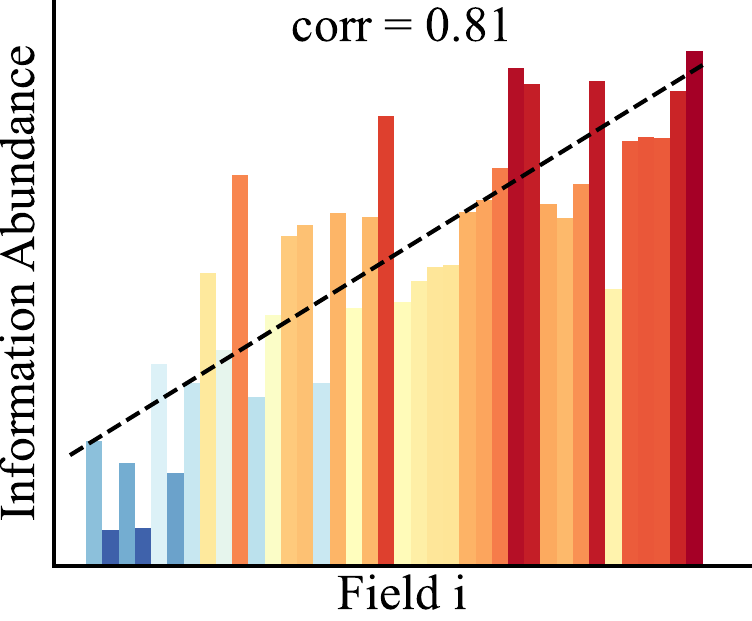}}%
            \end{minipage}%
            \begin{minipage}{0.3333\textwidth}
                \centering
                \subfloat[$\sum_{i=1}^N\mathrm{IA}(\mE_i^{\to j})$.\label{subfig:evd2f}]{\includegraphics[width=0.95\linewidth]{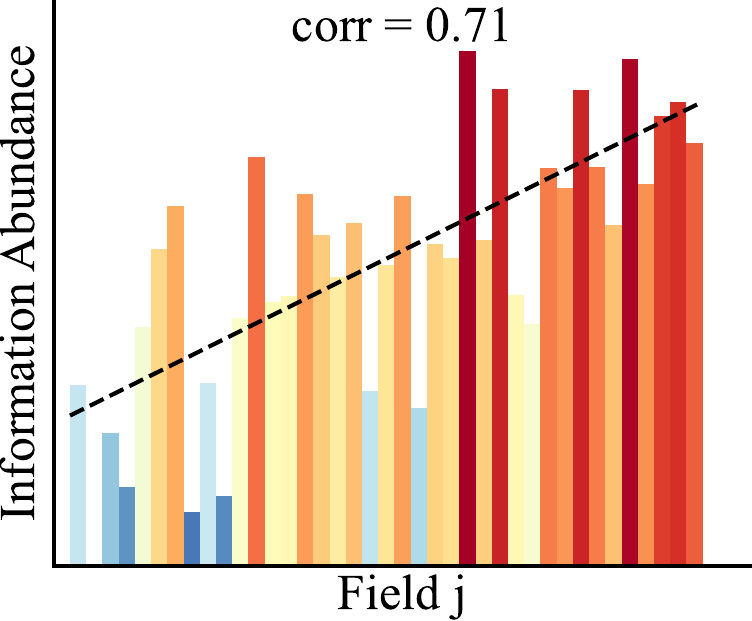}}%
            \end{minipage}%
            \caption{Information abundance of sub-embedding matrices for DCNv2, with field indices sorted by information abundance of corresponding raw embedding matrices. Higher or warmer indicates larger. It is observed that $\mathrm{IA}(\mE_i^{\to j})$ are co-influenced by both $\mathrm{IA}(\mE_i)$ and $\mathrm{IA}(\mE_j)$.}
        \end{minipage}%
    \end{minipage}%
    \begin{minipage}{0.28\textwidth}
        \centering
        \begin{minipage}{0.95\textwidth}
            \centering
            \includegraphics[width=0.95\linewidth]{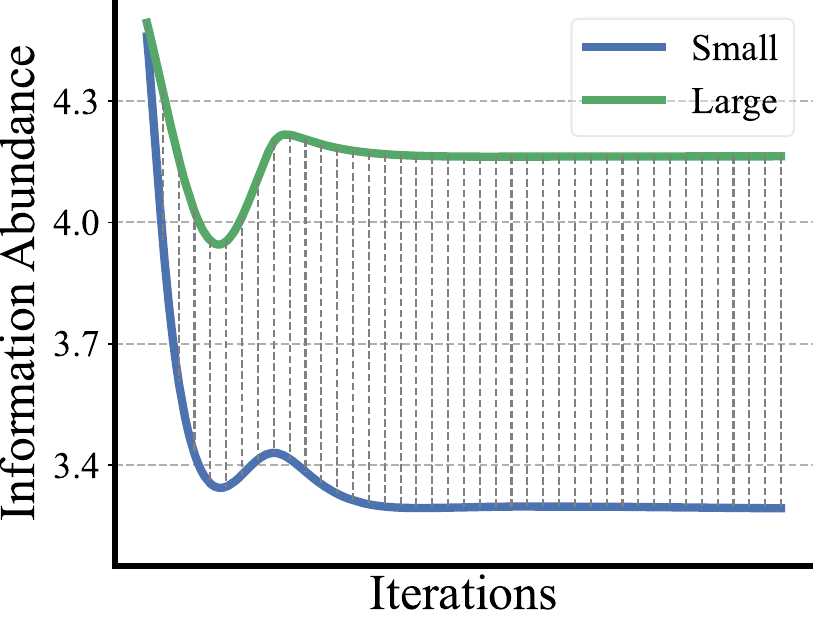}
            \caption{$\mathrm{IA}(\mE_1)$ for toy experiments. ``Small'' and ``Large'' refers to the cardinality of $\mathcal{X}_3$.}
            \label{fig:toy}
        \end{minipage}
    \end{minipage}
\end{figure*}

\paragraph{Evidence I: Empirical analysis on models with sub-embeddings.} DCNv2~\citep{dcnv2} incorporates a crossing network parameterized with transformation matrices $\mW_{i\to j}$~\citep{fmfm} over each field pair to project an embedding vector from field $i$ before interaction with field $j$. By collecting all projected embedding vectors, DCNv2 can be regarded to implicitly generate field-aware sub-embeddings $\mE_i^{\to 1},\mE_i^{\to 2},...,\mE_i^{\to N}$ to interact with all fields from embedding matrix $\mE_i$, \lms{using}
\[
\mE_i^{\to j}=\mE_i\mW_{i\to j}^\top.
\]
DCNv2 consists of multiple stacked cross layers, and we only discuss the first layer as simplification. To determine the collapse of sub-embedding matrices, we calculate $\mathrm{IA}(\mE_i^{\to j})$ for all $i,j$ pair and show them in Figure~\ref{subfig:evd2d}. For convenience, we pre-sort the field indices by the ascending order of information abundance, \textit{i.e.}, $i$ is ordered according to $\mathrm{IA}(\mE_i)$, similar to $j$. We can observe that $\mathrm{IA}(\mE_i^{\to j})$ is approximately increasing along $i$, which is trivial since $\mE_i^{\to j}$ is simply a projection of $\mE_i$. Interestingly, another correlation can be observed that the information abundance of sub-embeddings is co-influenced by the fields it interacts with, reflected by the increasing trend along $j$, especially for those with larger $i$. For instance, we further calculate the summation of $\mathrm{IA}(\mE_i^{\to j})$ over $j$ or $i$ to study the effect of the other single variable, shown in Figure~\ref{subfig:evd2e} and Figure~\ref{subfig:evd2f}. The increasing trend and the correlation factor \lms{confirm} the co-influence of $i$ and $j$. We \lms{further} analyze the IA of  FFM~\citep{ffm} model which also consists of sub-embeddings as DCNv2, obtaining similar observations shown in Appendx~\ref{apdx:ffm}.


\paragraph{Evidence II: Theoretical analysis on general recommendation models.}We now theoretically present how collapse is caused by feature interaction in general models, even without sub-embedding. For simplicity, we consider an FM-style~\citep{fm} feature interaction. Formally, the interaction process is defined by
\[
    h=\sum_{i=1}^N\sum_{j=1}^{i-1}\ve_i^\top\ve_j=\sum_{i=1}^N\sum_{j=1}^{i-1}\1_{x_i}^\top \mE_i\mE_j^\top \1_{x_j},
\]
where $h$ is the combined feature as mentioned before. Without loss of generality, we discuss one specific row $\ve_1$ of $\mE_1$ and keep other embedding matrices fixed. Consider a minibatch with batch size $B$. Denote $\sigma_{i,k}$ as the $k$-th singular value of $\mE_i$, and denote $\vu_{i,k}$, $\vv_{i,k}$ as corresponding singular vectors. We have

\begin{align*}
    \frac{\partial\mathcal{L}}{\partial \ve_1}
    &=\frac1B\sum_{b=1}^B\frac{\partial\ell^{(b)}}{\partial h^{(b)}}\cdot\frac{\partial h^{(b)}}{\partial\ve_1}
    =\frac1B\sum_{b=1}^B\frac{\partial\ell^{(b)}}{\partial h^{(b)}}\cdot\sum_{i=2}^N \mE_i^\top\1_{x_i^{(b)}} \\
    &=\frac1B\sum_{b=1}^B\frac{\partial\ell^{(b)}}{\partial h^{(b)}}\cdot\sum_{i=2}^N \sum_{k=1}^K\sigma_{i,k}\vv_{i,k}\vu_{i,k}^\top\1_{x_i^{(b)}} \\
    &=\sum_{i=2}^N\sum_{k=1}^K\left(\frac1B\sum_{b=1}^B\frac{\partial\ell^{(b)}}{\partial h^{(b)}}\vu_{i,k}^\top\1_{x_i^{(b)}}\right)\sigma_{i,k}\vv_{i,k} \\
    &=\sum_{i=2}^N\sum_{k=1}^K \alpha_{i,k}\sigma_{i,k}\vv_{i,k}=\sum_{i=2}^N\vtheta_i, \\
    &\text{where}\quad\vtheta_i=\sum_{k=1}^K\alpha_{i,k}\sigma_{i,k}\vv_{i,k}.
\end{align*}

The equation means that the gradient can be decomposed into field-specific terms. We consider the component $\vtheta_i$ for a certain field $i$, which is further decomposed into \lms{spectra} for the corresponding embedding matrix $\mE_i$. From the form of $\vtheta_i$, it is observed that $\{\alpha_{i,k}\}$ are $\bm{\sigma}_i$-agnostic scalars determined by the training data and objective function. Thus, the variety of $\bm{\sigma}_i$ significantly influences the composition of $\vtheta_i$. For those larger $\sigma_{i,k}$, the gradient component $\vtheta_i$ will be weighted more heavily along the corresponding \lms{spectra} $\vv_{i,k}$. When $\mE_i$ is low-information-abundance, the components of $\vtheta_i$ weigh imbalancely, resulting in the degeneration of $\ve_1$. Since different $\ve_1$ affects only $\alpha_{i,k}$ instead of $\sigma_{i,k}$ and $\vv_{i,k}$, all rows of $\mE_1$ degenerates in similar manners and finally form a collapsed matrix.

To further illustrate, we conduct a toy experiment over synthetic data. Suppose there are $N=3$ fields, and we set $D_3$ to different values with $D_3<K$ and $D_3\gg K$ to simulate low-information-abundance and high-information-abundance cases, which matches the diverse range of the field cardinality in real-world scenarios. We train $\mE_1$ while keeping $\mE_2,\mE_3$ fixed. Details of experiment setups are discussed in Appendix~\ref{apdx:toy}. We show the information abundance of $\mE_1$ along the training process for the two cases in Figure~\ref{fig:toy}. It is observed that interacting with a low-information-abundance matrix will result in a collapsed embedding.

\paragraph{Summary: How is collapse caused in recommendation models?} Evidence I highlights that interacting with a low-information-abundance field will result in a more collapsed sub-embedding. By considering the fact that sub-embeddings reflect the effect when fields interact since it originates from raw embeddings, we recognize the inherent mechanism of feature interaction to cause collapse, which is further confirmed by our theoretical analysis. We conclude the \emph{interaction-collapse theory}:

\begin{tcolorbox}[colback=blue!2!white,leftrule=2.5mm,size=title]
    \emph{Finding 1 (Interaction-Collapse Theory). In feature interaction of recommendation models, fields with low-information-abundance embeddings constrain the information abundance of other fields, resulting in collapsed embedding matrices.}
\end{tcolorbox}

The interaction-collapse theory generally suggests that feature interaction is the primary catalyst for collapse, thereby imposing constraints on the ideal scalability.

\subsection{Is It Sufficient to Avoid Collapse for Scalability?}

\label{subsec:generalizability}

Following our discussion above, we have shown that the feature interaction process of recommendation models leads to collapse and thus limits the model scalability. We now discuss its \emph{negative proposition}, \textit{i.e.}, whether suppressing the feature interaction to mitigate collapse leads to model scalability. To answer this question, we design the following two experiments to compare standard models and models with feature interaction suppressed. 

\begin{figure*}
    \centering
    \begin{minipage}{0.6\textwidth}
        \centering
        \begin{minipage}{0.95\textwidth}
            \centering
            \begin{minipage}{0.3333\textwidth}
                \centering
                \subfloat[IA w/ 10x size.\label{subfig:evd3}]{\includegraphics[width=0.95\textwidth]{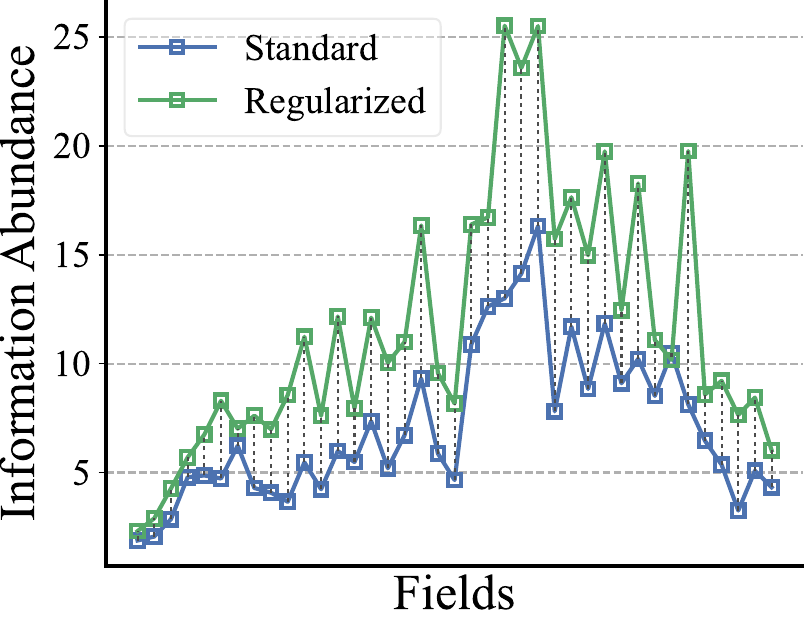}}%
            \end{minipage}%
            \begin{minipage}{0.3333\textwidth}
                \centering
                \subfloat[Test AUC w.r.t. size.]{\includegraphics[width=0.95\textwidth]{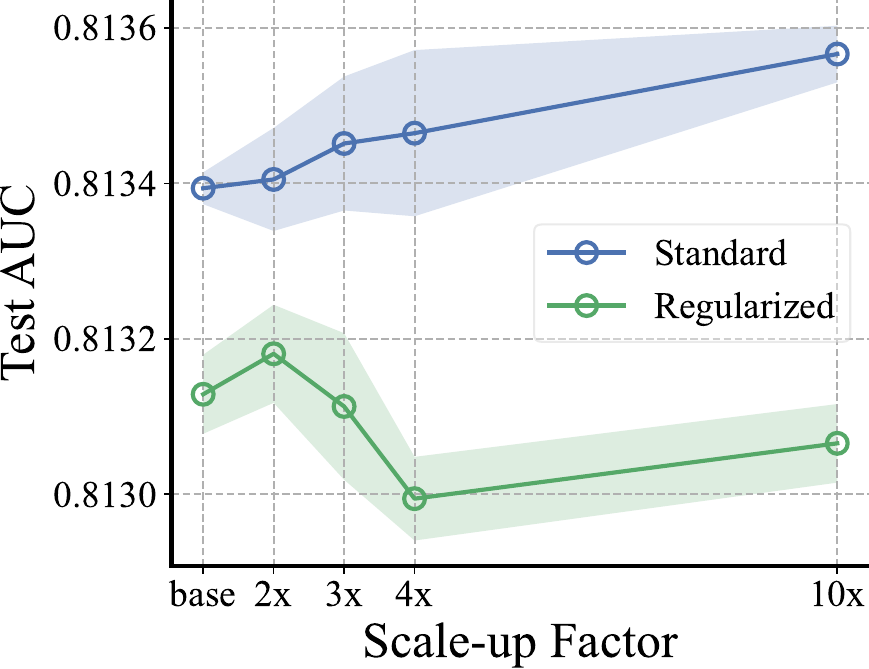}}%
            \end{minipage}%
            \begin{minipage}{0.3333\textwidth}
                \centering
                \subfloat[Training curve.]{\includegraphics[width=0.95\textwidth]{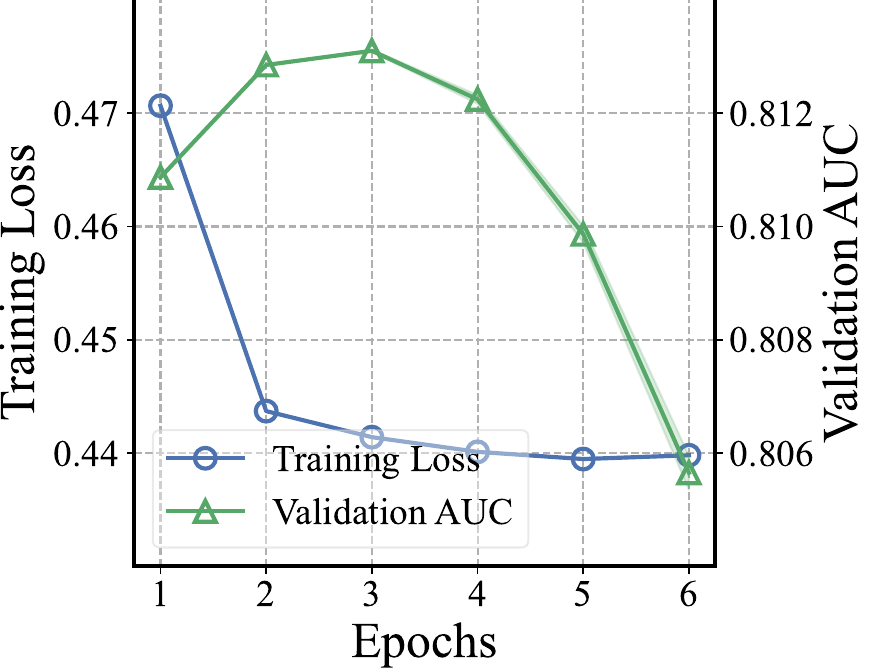}}%
            \end{minipage}
            \caption{Experimental results of Evidence III. Restricting DCNv2 leads to higher information abundance, yet the model suffers from over-fitting, thus resulting in non-scalability.}
            \label{fig:evd3}
        \end{minipage}
    \end{minipage}%
    \begin{minipage}{0.4\textwidth}
        \centering
        \begin{minipage}{0.95\textwidth}
            \centering
            \begin{minipage}{0.495\linewidth}
                \centering
                \subfloat[IA w/ 10x size.]{\includegraphics[width=0.95\textwidth]{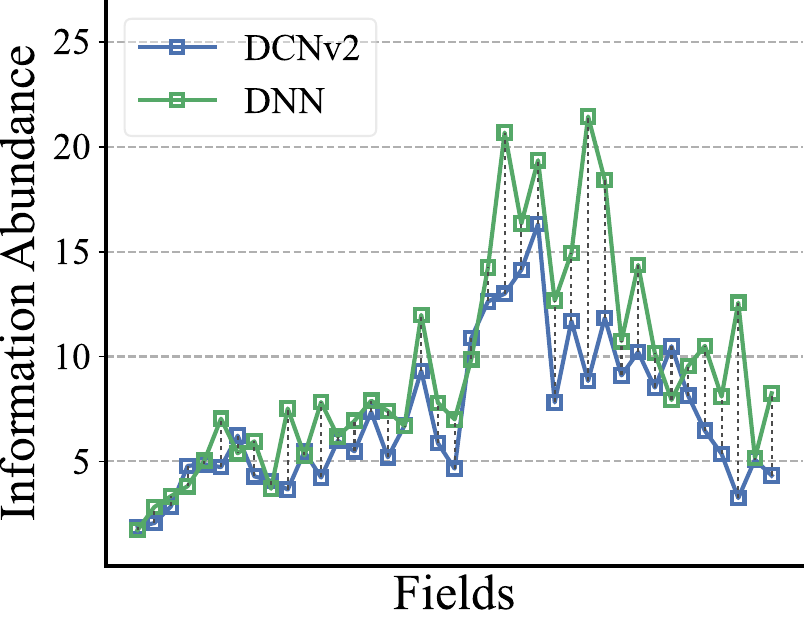}}
            \end{minipage}%
            \begin{minipage}{0.505\linewidth}
                \centering
                \subfloat[Test AUC w.r.t. size.]{\includegraphics[width=0.95\textwidth]{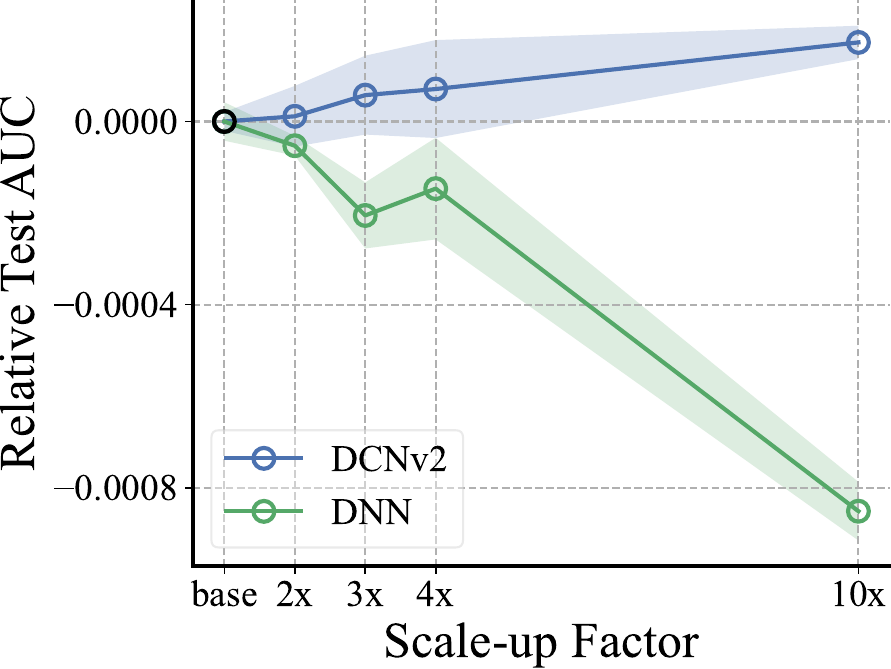}}
            \end{minipage}%
            \caption{Experimental results of Evidence IV. Despite higher information abundance, the performance of DNN drops w.r.t. model size.}
            \label{fig:evd4}
        \end{minipage}
    \end{minipage}
\end{figure*}

\paragraph{Evidence III: Limiting the modules in interaction that leads to collapse.} Evidence I shows that a projection $\mW_{i\to j}$ is learned to adjust information abundance for sub-embeddings and lead to collapse. We now inverstigate how surpressing such effect would result in model scalability by introducing the following regularization with learnable parameter $\lambda_{ij}$:
\[
    \ell_{reg}=\sum_{i=1}^N\sum_{j=1}^N\left\|\mW_{i\to j}^\top \mW_{i\to j}-\lambda_{ij} \mI\right\|_{\mathrm{F}}^2,
\]

\lms{which} regularizes the projection matrix to be a multiplication of an unitary matrix. This way, $\mW_{i\to j}$ will preserve all normalized singular values and maintain the information abundance after projection. We experiment with various embedding sizes and compare the changes in performance, the information abundances, and the optimization dynamics for standard and regularized models. Results are shown in Figure~\ref{fig:evd3}. As anticipated, regularization in DCNv2 helps learn embeddings with higher information abundance. Nevertheless, the model presents unexpected results whereby the scalability does not improve or worsen even if the collapse is alleviated, and it is found that such a model overfits during the learning process with the training loss consistently decreasing and the validation AUC dropping.



\paragraph{Evidence IV: Directly avoiding explicit interaction.} We now investigate how directly suppressing the feature interaction would affect the scalability. We discuss DNN, which consists of a plain interaction module that concatenates all feature vectors from different fields and processes them with an MLP. Since DNN does not conduct explicit 2-order feature interaction~\citep{rendle2020neural}, it would suffer less from collapse following our previous interaction-collapse theory. We compare the learned embeddings of DCNv2 and DNN and their performance with the growth of embedding size. Considering that different architectures or objectives may differ in modeling, we mainly discuss the performance trend as a fair comparison. Results are shown in Figure~\ref{fig:evd4}. DNN learns less-collapsed embedding matrices, reflected by higher information abundance than DCNv2. Yet perversely, the AUC of DNN drops when increasing the embedding size. Such observations show that DNN falls into the issue of overfitting and lacks scalability, even though it suffers less from collapse.

\paragraph{Summary: Does suppressing collapse definitely improve scalability?} Regularized DCNv2 and DNN are both models with feature interaction suppressed, and they learn less-collapsed embedding matrices than DCNv2, as expected. Yet observations in evidence III\&IV demonstrate that regularized DCNv2 and DNN both \lms{do not scale in AUC} with the growth of model size and suffer from serious overfitting. We conclude the following finding:

\begin{tcolorbox}[colback=blue!2!white,leftrule=2.5mm,size=title]
    \emph{%
        Finding 2. A less-collapsed model with feature interaction suppressed \lms{improperly} is insufficient for scalability due to overfitting concern.
    }
\end{tcolorbox}

Such a finding is plausible, considering that feature interaction brings domain knowledge of higher-order correlations in recommender systems and helps form generalizable representations. When feature interaction is suppressed, models tend to fit noise as the embedding size increases, resulting in reduced generalization.

\section{\lms{Multi-Embedding} Design}

In this section, we present a simple \emph{multi-embedding} design, which serves as an effective scaling \lms{mechanism} applicable to a wide range of \lms{recommendation} model architectures. We introduce the overall architecture, present experimental results, and analyze how multi-embedding works. We also discuss the role of data to give a comprehensive analysys for multi-embedding.

\subsection{Multi-Embedding}
\label{subsec:me-arch}

\begin{figure*}[ht]
    \centering
    \begin{minipage}{\textwidth}
        \centering
        \includegraphics[width=\textwidth]{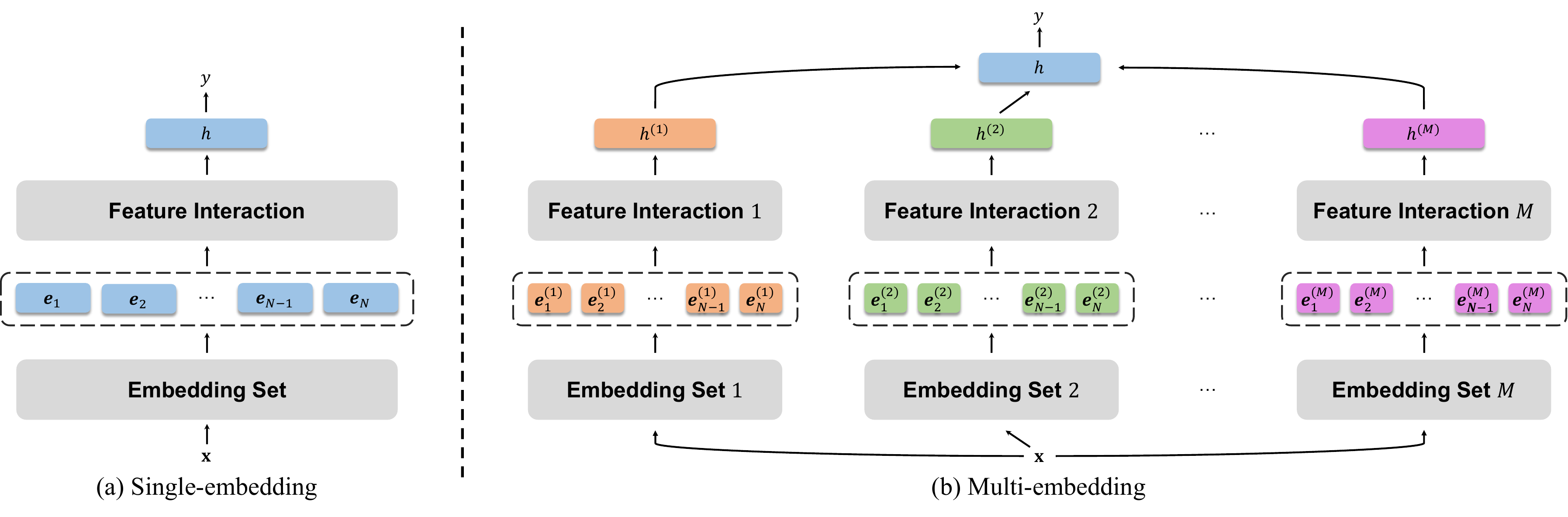}
        \caption{Architectures of single-embedding (left) and multi-embedding (right) models.}
        \label{fig:exp-me-arch}
    \end{minipage}%
\end{figure*}

The two-sided effect of feature interaction for scalability implies a \emph{principle} for model design. That is, a scalable model should be capable of less-collapsed embeddings within the existing feature interaction framework instead of removing interaction. Based on this principle, we propose \emph{multi-embedding} or \emph{ME} as a simple yet efficient design to improve scalability. Specifically, we scale up the number of independent and complete embedding sets instead of the embedding size, and incorporate embedding-set-specific feature interaction modules. \lms{In line with} previous works such as group convolution~\citep{alexnet}, multi-head attention~\citep{transformer}, and other decoupling-based works in recommender systems~\citep{liu2019single,liu2022disentangled,weston2013nonlinear}, such design allows the model to learn different interaction patterns jointly and result in embedding sets with large diversity, while a single-embedding model would be limited in pattern \lms{extraction} and suffer from severe collapse. With multi-embedding, the model is less influenced by the interaction-collapse theory and mitigates embedding collapse while keeping the original interaction modules. Formally, a \lms{recommendation} model with $M$ sets of embeddings is defined as

\vspace{-20pt}
\begin{align*}
    \ve_i^{(m)}&=\left(\mE_i^{(m)}\right)^{\top}\1_{x_i},\ \forall i\in\{1,2,...,N\}, \\
    h^{(m)}&=I^{(m)}\left(\ve_1^{(m)},\ve_2^{(m)},...,\ve_N^{(m)}\right), \\
    h&=\frac1M\sum_{m=1}^M h^{(m)},\quad \hat{y}=F(h),
\end{align*}

\vspace{-10pt}
where $m$ stands for the index of embedding set. One requirement of multi-embedding is that there should be non-linearities such as ReLU within the interaction \lms{module} $I$; otherwise, the model is equivalent to single-embedding and hence does not capture different patterns.\footnote{See Appendix~\ref{apdx:non-linearity}.} As a solution, we add a non-linear projection after interaction for the models with linear interaction modules and reduce one MLP layer for \lms{postprocessing module} $F$ to achieve a fair comparison. An overall architecture comparison of single-embedding and mult-embedding models is shown in Figure~\ref{fig:exp-me-arch}.

\subsection{Experiments}

\paragraph{Setup.} We conduct our experiments on two datasets for recommender systems: Criteo~\citep{criteo} and Avazu~\citep{avazu}, which are large and \lms{challenging} benchmark datasets widely used in recommender systems. We experiment on baseline models including DNN, IPNN~\citep{ipnn}, NFwFM~\citep{fwfm}, xDeepFM~\citep{xdeepfm}, DCNv2~\citep{dcnv2}, FinalMLP~\citep{finalmlp} and their corresponding multi-embedding variants with 2x, 3x, 4x and 10x model size. Here NFwFM is a variant of NFM~\cite{nfm} by replacing FM with FwFM. All experiments are performed with 8/1/1 training/validation/test splits, and we apply early stopping based on validation AUC. More details are shown in Appendix~\ref{apdx:exp-settings}.

\begin{table*}[ht]
    \caption{Test AUC for different models. Higher indicates better. Underlined and bolded values refer to the best performance with single-embedding (SE) and multi-embedding (ME), respectively.}
    \label{tab:main-exp}
    \begin{center}
        \resizebox{\textwidth}{!}{%
            \begin{tabular}{cccccccccccc}
                \toprule
                \multicolumn{2}{c}{\multirow{2.5}{*}{\large Model}}    & \multicolumn{5}{c}{Criteo}                                       & \multicolumn{5}{c}{Avazu}                                        \\ \cmidrule(lr){3-7} \cmidrule(lr){8-12}
                                          &                         & base                       & 2x      & 3x      & 4x      & 10x     & base                       & 2x      & 3x      & 4x      & 10x     \\ \midrule
                \multirow{2}{*}{DNN}      & SE                  & \multirow{2}{*}{\underline{0.81228}} & \underline{0.81222} & 0.81207 & 0.81213 & 0.81142 & \multirow{2}{*}{0.78744} & \underline{0.78759} & 0.78752 & 0.78728 & 0.78648 \\
                                          & ME                &                          & 0.81261 & \textbf{0.81288} & \textbf{0.81289} & \textbf{0.81287} &                          & 0.78805 & 0.78826 & 0.78862 & \textbf{0.78884} \\ \midrule
                \multirow{2}{*}{IPNN}     & SE                  & \multirow{2}{*}{\underline{0.81272}} & \underline{0.81273} & \underline{0.81272} & \underline{0.81271} & 0.81262 & \multirow{2}{*}{0.78732} & \underline{0.78741} & 0.78738 & \underline{0.78750} & \underline{0.78745} \\
                                          & ME                &                          & 0.81268 & 0.81270 & 0.81273 & \textbf{0.81311} &                          & 0.78806 & 0.78868 & 0.78902 & \textbf{0.78949} \\ \midrule
                \multirow{2}{*}{NFwFM}    & SE                  & \multirow{2}{*}{0.81059} & 0.81087 & 0.81090 & \underline{0.81112} & \underline{0.81113} & \multirow{2}{*}{0.78684} & 0.78757 & 0.78783 & \underline{0.78794} &    \underline{0.78799}   \\
                                          & ME                &                          & 0.81128 & 0.81153 & 0.81171 & \textbf{0.81210} &                          & 0.78868 & 0.78901 & 0.78932 &    \textbf{0.78974}   \\ \midrule
                \multirow{2}{*}{xDeepFM}  & SE                  & \multirow{2}{*}{\underline{0.81217}} & 0.81180 & 0.81167 & 0.81137 & 0.81116 & \multirow{2}{*}{\underline{0.78743}} & \underline{0.78750} & 0.78714 & 0.78735 & 0.78693 \\
                                          & ME                &                          & 0.81236 & 0.81239 & 0.81255 & \textbf{0.81299} &                          & 0.78848 & 0.78886 & 0.78894 & \textbf{0.78927} \\ \midrule
                \multirow{2}{*}{DCNv2}    & SE                  & \multirow{2}{*}{0.81339} & 0.81341 & 0.81345 & 0.81346 & \underline{0.81357} & \multirow{2}{*}{0.78786} & 0.78835 & \underline{0.78854} & \underline{0.78852} & \underline{0.78856} \\
                                          & ME                &                          & 0.81348 & 0.81361 & \textbf{0.81382} & \textbf{0.81385} &                          & 0.78862 & 0.78882 & 0.78907 & \textbf{0.78942} \\ \midrule
                \multirow{2}{*}{FinalMLP} & SE                  & \multirow{2}{*}{\underline{0.81259}} & \underline{0.81262} & 0.81248 & 0.81240 & 0.81175 & \multirow{2}{*}{0.78751} & \underline{0.78797} & \underline{0.78795} & 0.78742 & 0.78662 \\
                                          & ME                &                          & 0.81290 & \textbf{0.81302} & \textbf{0.81303} & \textbf{0.81303} &                          & 0.78821  & \textbf{0.78831} & \textbf{0.78836} & \textbf{0.78830} \\ \bottomrule
            \end{tabular}
        }
    \end{center}
\end{table*}

\paragraph{Results.} We repeat each experiment 3 times and report the average test AUC with different scaling factors of the model size. Results are shown in Table~\ref{tab:main-exp}. For the experiments with single-embedding, we observe that all the models demonstrate poor scalability. Only DCNv2 and NFwFM show slight improvements with increasing embedding sizes, with gains of 0.00036 on Criteo and 0.00093 on Avazu, respectively. For DNN, xDeepFM, and FinalMLP, which rely highly on non-explicit interaction, the performance even drops (0.00136 on Criteo and 0.00118 on Avazu) when scaled up to 10x, as discussed in \Secref{subsec:generalizability}. In contrast to single-embedding, our multi-embedding shows consistent and remarkable improvement with the growth of the embedding size, and the highest performance is always achieved with the largest 10x size. For DCNv2 and NFwFM, multi-embedding gains 0.00099 on Critio and 0.00223 on Avazu by scaling up to 10x, which is never obtained by single-embedding. Over all models and datasets, compared with baselines, the largest models averagely achieve 0.00110 improvement on the test AUC\footnote{A slightly higher AUC at 0.001-level is regarded significant~\citep{wide_and_deep,deepfm,autoint,eulernet}}. Multi-embedding provides a methodology to break through the non-scalability limit of existing models. We visualize the scalability of multi-embedding on Criteo dataset in Figure~\ref{subfig:scalability-criteo}. The standard deviation and detailed scalability comparison are shown in Appendix~\ref{apdx:exp-results}. 


\begin{figure*}[ht]
    \centering
    \begin{minipage}{0.5\textwidth}
        \centering
        \begin{minipage}{0.95\textwidth}
            \centering
            \begin{minipage}{0.5\linewidth}%
                \centering
                \subfloat[\label{subfig:scalability-criteo}Multi-embedding on Criteo.]{\includegraphics[width=0.95\textwidth]{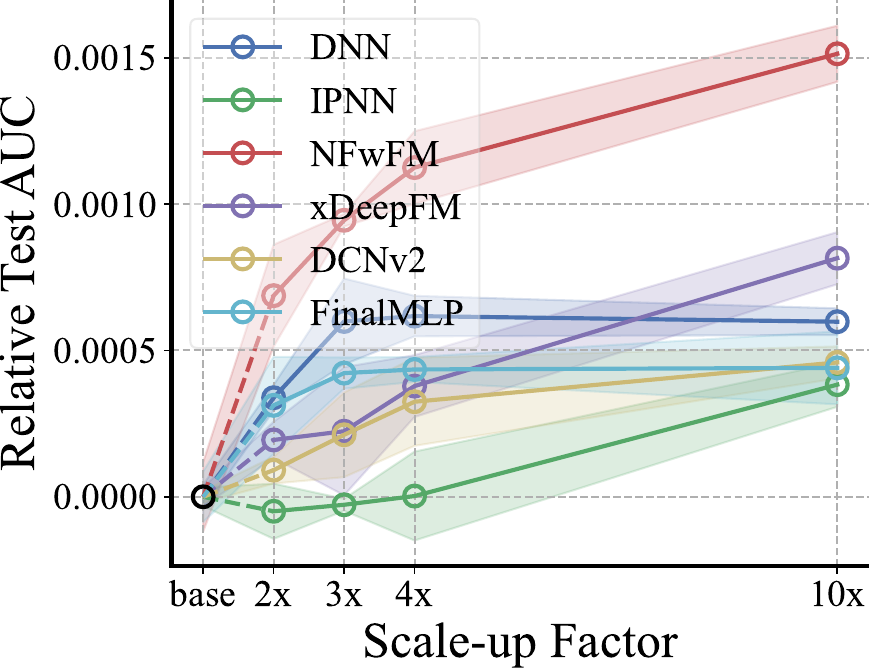}}
            \end{minipage}%
            \begin{minipage}{0.5\linewidth}%
                \centering
                \subfloat[\label{subfig:ia-single-vs-multiple}$\mathrm{IA}(\mE_i)$ on DCNv2.]{\includegraphics[width=0.95\textwidth]{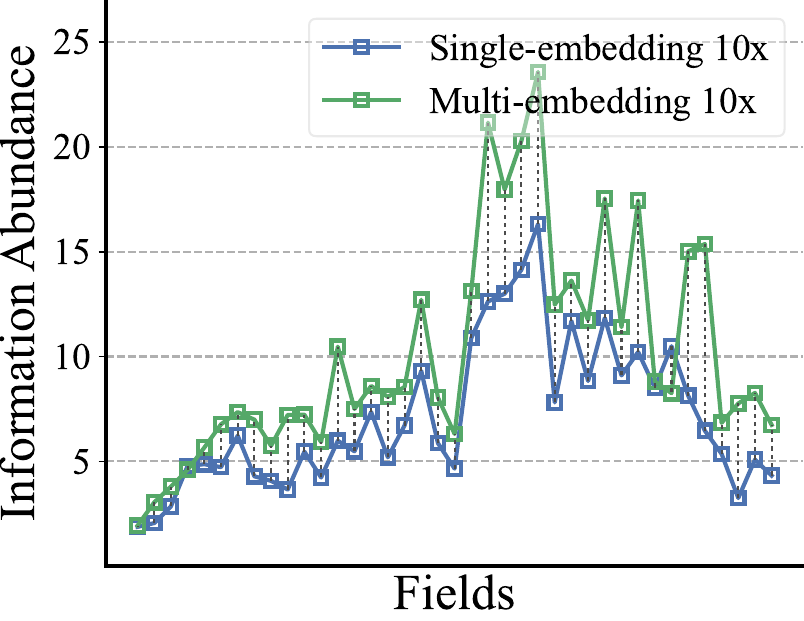}}
            \end{minipage}%
            \caption{Effectiveness of multi-embedding. \textbf{(a)} Model performance consistently improves with increasing size when using multi-embedding on 6 mainstream models. \textbf{(b)} Multi-embedding consistently enhances information abundance across all fields compared with single-embedding.}
            \label{fig:me-effectiveness}
        \end{minipage}%
    \end{minipage}%
    \begin{minipage}{0.5\textwidth}
        \centering
        \begin{minipage}{0.95\textwidth}
            \centering
            \begin{minipage}{0.5\textwidth}
                \centering
                \subfloat[$\sum_{i=1}^N\mathrm{IA}(\mE_i^{\to j})$, SE.\label{subfig:evd2apdxd}]{\includegraphics[width=0.89\textwidth]{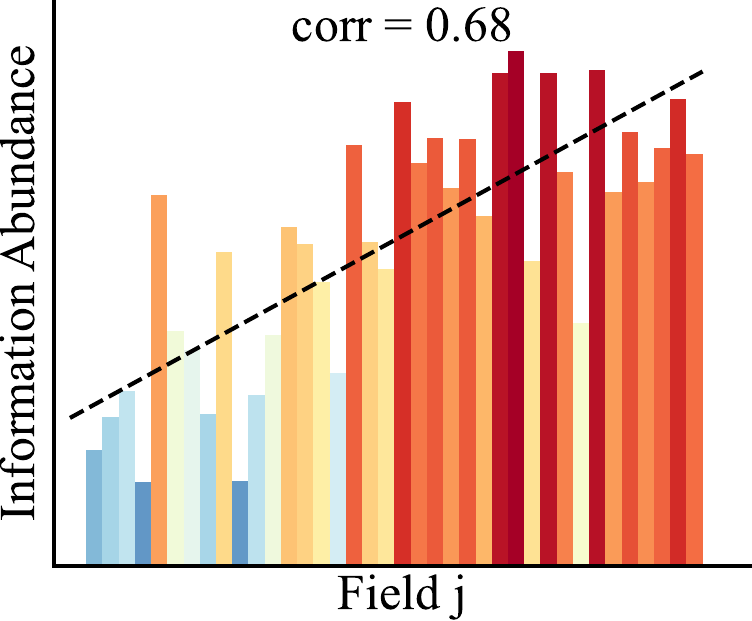}}%
            \end{minipage}%
            \begin{minipage}{0.5\textwidth}
                \centering
                \subfloat[$\sum_{i=1}^N\mathrm{IA}(\mE_i^{\to j})$, ME.\label{subfig:evd2apdxc}]{\includegraphics[width=0.89\textwidth]{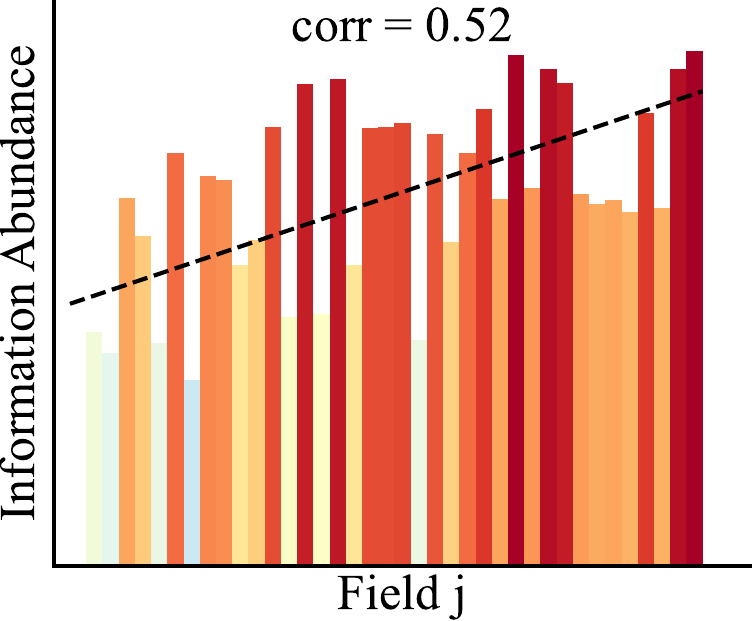}}%
            \end{minipage}
            \caption{Information abundance of sub-embedding matrices for single-embedding and multi-embedding on DCNv2, with field indices sorted by information abundance of corresponding raw embedding matrices. Higher or warmer indicates larger. $\mathrm{IA}(\mE_i^{\to j})$ are less influenced by $j$ or $\mathrm{IA}(\mE_j)$ in multi-embedding.}
            \label{fig:evd2apdx}
        \end{minipage}
    \end{minipage}
\end{figure*}

\paragraph{Mitigation of embedding collapse.} We compare the information abundance of single-embedding and multi-embedding DCNv2 with the largest 10x embedding size to measure the mitigation of collapse, with all embedding sets for a single field concatenated together as the overall embedding in multi-embedding DCNv2. Results are shown in Figure~\ref{subfig:ia-single-vs-multiple}. It is observed that multi-embedding DCNv2 consistently increases the information abundance for all fields compared with single-embedding DCNv2, especially for fields with larger cardinality. These results indicate that multi-embedding is a simple yet effective method to mitigate embedding collapse for scalability gain without introducing significant computational resources or hyper-parameters.

\paragraph{Deployment in the online system.} After the online A/B testing in January 2023, the multi-embedding paradigm has been successfully deployed in Tencent's Online Advertising Platform, one of the largest advertisement recommendation systems. Upgrading the click prediction model from the single-embedding paradigm to our proposed multi-embedding paradigm in WeChat Moments leads to a 3.9\% GMV (Gross Merchandise Value) lift, which brings hundreds of millions of dollars in revenue lift per year.
Details are introduced in \citet{pan2024ad}.

\subsection{How Multi-Embedding works?}
\label{sec:how_works}

\paragraph{Less influenced by the interaction-collapse theory.} Following our previous interaction-collapse theory and corresponding analysis, embedding collapse is caused by feature interaction between different fields, reflected by the co-influence on the information abundance of sub-embeddings. We show that multi-embedding suffers less from such an effect. Recall in Section~\ref{sec:fi} that we calculate $\sum_{i=1}^N\mathrm{IA}(\mE_i^{\to j})$ to compare how $\mathrm{IA}(\mE_i^{\to j})$ is influenced by the field to interact with. We here correspondingly visualize the result for multi-embedding and single-embedding DCNv2, shown in Figure~\ref{fig:evd2apdx}. It can be observed that the correlation factor in multi-embedding is significantly less than that in single-embedding (0.52 vs. 0.68). Thus, the information abundance is less influenced by the field it interacts with, suffering less from the impact of the interaction-collapse theory.


\begin{figure*}[ht]
    \centering
    \begin{minipage}{0.25\textwidth}
        \centering
        \subfloat[\label{subfig:corresponding-angle}Diversity of ME \& SE.]{\includegraphics[width=0.95\textwidth]{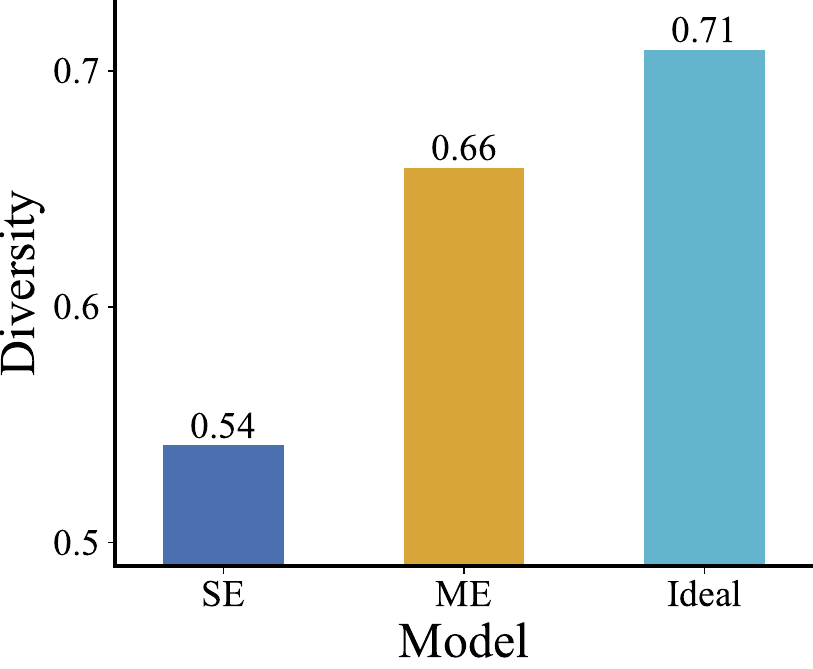}}
    \end{minipage}%
    \begin{minipage}{0.25\textwidth}
        \centering
        \subfloat[\label{subfig:vis-W}$\|\mW_{i\to j}^{(m)}\|_{\mathrm{F}}$.]{\includegraphics[width=0.9\textwidth]{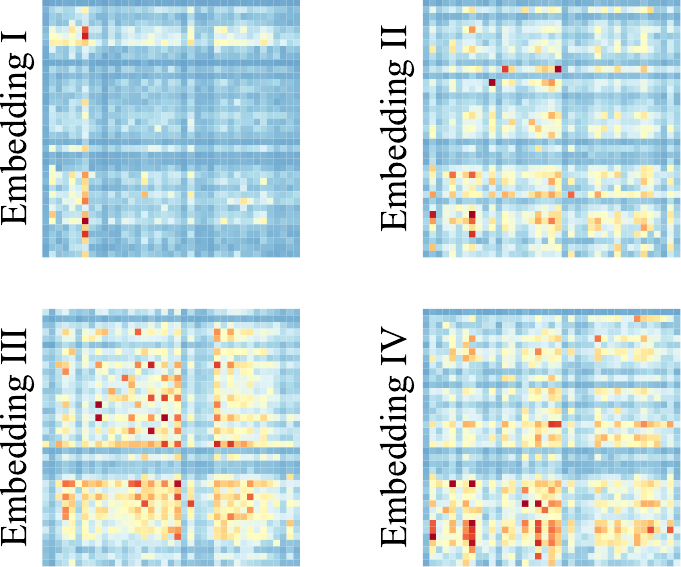}}
    \end{minipage}%
    \begin{minipage}{0.25\textwidth}
        \centering
        \subfloat[\label{subfig:weight-norm-align}Scaling up ME variants.]{\includegraphics[width=0.99\textwidth]{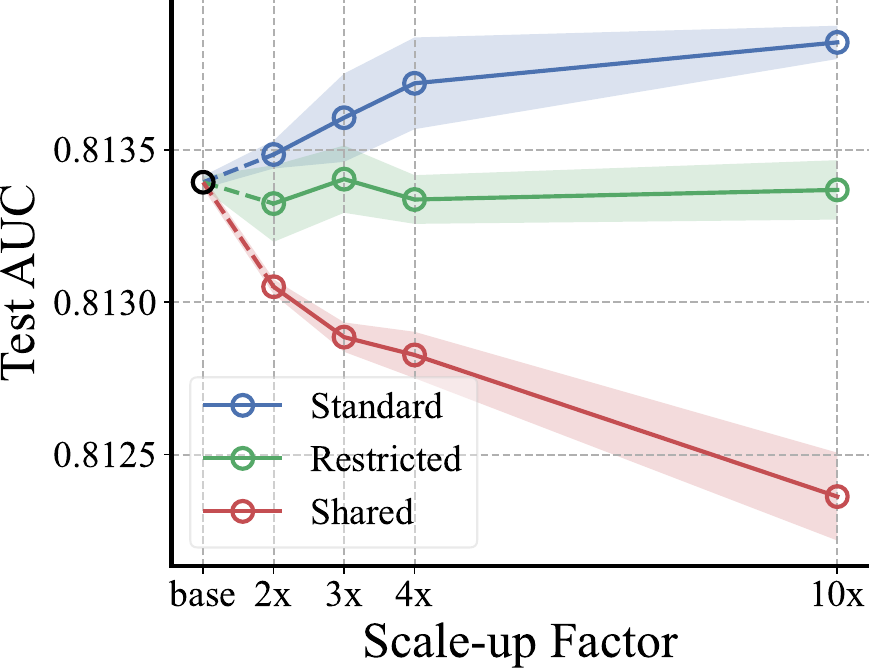}}
    \end{minipage}%
    \begin{minipage}{0.25\textwidth}
        \centering
        \subfloat[\label{subfig:weight-norm-align-bar}Diversity of ME variants.]{\includegraphics[width=0.95\textwidth]{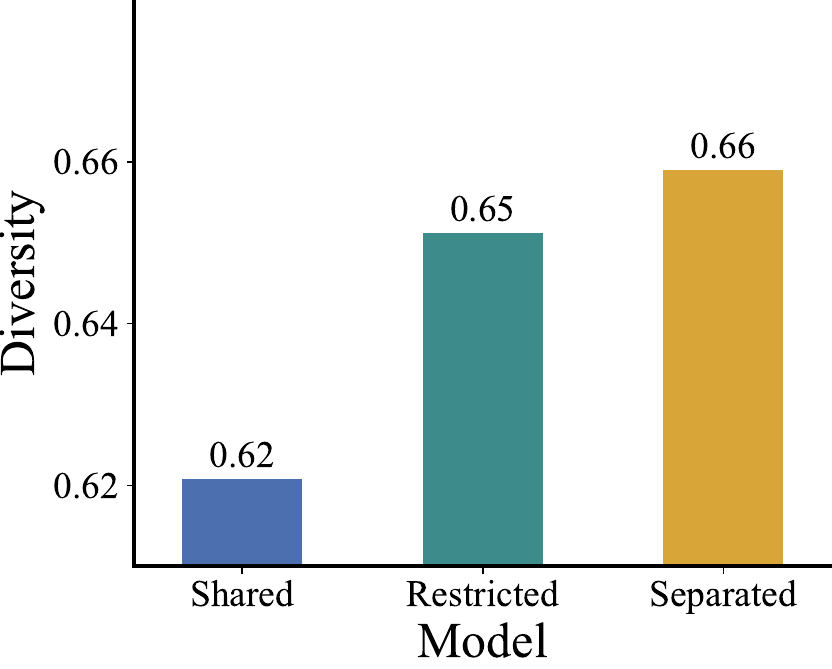}}
    \end{minipage}%
    \caption{Analysis of \lms{multi-embedding} (ME). \textbf{(a)}: Multi-embedding learns embedding sets with significantly larger diversity than single-embedding. \textbf{(b)}: Embedding-set-specific feature interaction modules capture different interaction patterns. \textbf{(c)}--\textbf{(d)}: Using variants of ME with non-separated interaction lead to worse scalability and lower embedding set diversity.}
    \label{fig:exp-me-analysis}
\end{figure*}

\paragraph{Mitigating collapse from embedding diversity.} 
We further demonstrate that multi-embedding mitigates collapse by allowing the diversity of embedding sets. To illustrate, we introduce the cosine of \emph{principal angle}~\citep{miao1992principal}, $\cos\left(\bm{\phi}_i^{m\leftrightarrow m'}\right)$, to measure space similarity between a pair of embedding sets $m,m'$ for any certain field $i$, calculated by the following further singular value decomposition:

\vspace{-10pt}
\[
\big(\mU_i^{(m)}\big)^\top\mU_i^{(m')}=\mP_i^{(m)}\mathrm{diag}\left(\cos\big(\bm{\phi}_i^{m\leftrightarrow m'}\big)\right)\big(\mP_i^{(m')}\big)^\top.
\]

\vspace{-10pt}
A low-rank $\big(\mU_i^{(m)}\big)^\top\mU_i^{(m')}$ implies a high-rank overall embedding $[\mE_i^{(m)},\mE_i^{(m')}]$.\footnote{See Appendix~\ref{apdx:diversity_theory}.} We therefore discuss a gereralized measurement


\[
    \mathrm{div}(\mE_i^{(m)},\mE_i^{(m')})=1-\frac1K\left\|\cos\left(\bm{\phi}_i^{m\leftrightarrow m'}\right)\right\|_1
\]

to describe the diversity of embedding sets. A larger diversity implies larger information abundance for the overall embedding or better mitigation of collapse. For comparision, we split the embedding of a single-embedding DCNv2 and an ideal random-initialized matrix into embedding sets, and compare with multi-embedding DCNv2. We show the average diversity across all embedding set pairs and all fields in Figure~\ref{subfig:corresponding-angle}. It is shown that multi-embedding can significantly lower the embedding set similarity compared with single-embedding, mitigating embedding collapse.

\paragraph{Yielding diversity from separated interaction.} We further demonstrate the embedding diversity of multi-embedding models originates from embedding-set-specific feature interaction modules, which allows the embedding sets to capture diverse interaction patterns. On the one hand, we visualize $\|\mW_{i\to j}^{(m)}\|_{\mathrm{F}}$ as the interaction pattern~\citep{dcnv2} for a multi-embedding DCNv2 model in Figure~\ref{subfig:vis-W}. It is shown that the interaction modules learn various patterns. On the other hand, we compare multi-embedding with two variants with non-separated interaction: (a) All feature interaction modules are shared across all embedding sets, and (b) The divergence of $\|\mW_{i\to j}^{(m)}\|_{\mathrm{F}}$ across all embedding sets are restricted by regularizations. Results are shown in Figure~\ref{subfig:weight-norm-align} and Figure~\ref{subfig:weight-norm-align-bar}. Compared with the separated design in multi-embedding, the two variants of feature interaction design show worse scalability and embedding diversity, indicating that multi-embedding works from the separation of interaction modules.

\begin{table*}[ht]
\centering
\caption{Averaged information abundance under various embedding sizes and data amount.}
\label{tab:ia-various-data}
\vskip 5pt
\begin{tabular}{l|cccccccc|c}
    \toprule
    Embedding Size & 5               & 10              & 15              & 20              & 25              & 30              & 35              & 40              & 40, ME  \\ \midrule
    1\% data       & 3.74          & 5.96          & 7.97          & 9.23          & 10.44         & 11.51         & 12.39         & 13.30         & 14.99 \\
    3\% data       & 3.16          & 4.80          & 5.96          & 6.67          & 7.30          & 8.10          & 8.46          & 8.84          & 10.59 \\
    10\% data      & \textbf{2.96} & 4.28          & 5.08          & 5.73          & 6.43          & 6.79          & 7.11          & 7.51          & 8.63  \\
    30\% data      & 2.97          & \textbf{4.09} & \textbf{4.89} & \textbf{5.41} & \textbf{5.87} & 6.19          & 6.53          & 6.78          & 7.70  \\
    100\% data     & 3.29          & 4.73          & 5.41          & 5.70          & 5.95          & \textbf{6.19} & \textbf{6.48} & \textbf{6.64} & 7.64  \\ \bottomrule
    \end{tabular}
\end{table*}

\subsection{Role of Data in Embedding Collapse}
Throughout our work, we mainly focus on the model scalability and conclude the intrinsic issue, embedding collapse, of recommendation models. This considerable data amount of our benchmark datasets of experiments provides credibility to the embedding collapse phenomenon within a data-amount-agnostic manner. In this section we further discuss how the embedding collapse phenomenon acts under various data amount. To illustrate, we conduct additional experiments using variously sized subsets of the Criteo dataset. We measured the average information abundance in embedding matrices across different model scales, summarized in Table \ref{tab:ia-various-data}. From the results, it is observed that the data size can indeed affect the information abundance of embedding matrices, but the information abundance does not strictly increase or even decreases along with the data size, especially for larger models. Behind this finding is that, embedding collapse is determined by two aspects: (1) the data size, which increases the information abundance, and (2) the interaction-collapse law, which decreases the information abundance. Among all the results, only for experiments with $10\%\sim 100\%$ data size and embedding size 5, and $30\%\sim 100\%$ data size and embedding size 10, 15, 20, 25, can we observe the collapse is caused by the limited data. And for most of the other cases, the abnormal decreasing trend shows that the collapse is caused by the interaction-collapse law rather than the limited data. Also throughout the results, multi-embedding consistently outperforms single-embedding under various data amount, indicating the universality of our proposed multi-embedding design.


\section{Related Works}

\paragraph{Modules in recommender systems.} 
Plenty of existing works investigate the module design for recommender systems. A line of studies focuses on feature interaction process~\citep{mf, fm, ffm,ipnn, nfm, deepfm, fwfm, xdeepfm, autoint, afn, fmfm,dcnv2,finalmlp, eulernet}, which is specific for recommender systems. These works are built up to fuse domain-specific knowledge of recommender systems. In contrast to proposing new modules, our work starts from a view of machine learning and analyzes the existing models for scalability.

\paragraph{Collapse phenomenon.}
Neural collapse or representation collapse describes the degeneration of representation vectors with restricted variation. This phenomenon is widely studied in supervised learning~\citep{papyan2020prevalence,zhu2021geometric,tirer2022extended}, unsupervised contrastive learning~\citep{hua2021feature,jing2021understanding,gupta2022understanding}, transfer learning~\citep{aghajanyan2020better,kumar2022fine} and generative models~\citep{lsgan,sngan}. \citet{chi2022representation} discuss the representation collapse in sparse MoEs. Inspired by these works, we realize the embedding collapse of recommendation models when regarding embedding vectors as representations by their definition, yet we are facing the setting of field-level interaction, which has not previously been well studied.

\paragraph{Intrinsic dimensions and compression theories.}
To describe the complexity of data, existing works include intrinsic-dimension-based quantification~\citep{levina2004maximum,ansuini2019intrinsic,pope2020intrinsic} and pruning-based analysis~\citep{wen2017coordinating,alvarez2017compression,fmfm}. Our SVD-based concept of information abundance is related to these works.

\section{Conclusion}

In this paper, we highlight the non-scalability issue of existing recommendation models and identify the embedding collapse phenomenon that hinders scalability. From empirical and theoretical analysis around embedding collapse, we conclude the two-sided effect of feature interaction on scalability, \textit{i.e.}, feature interaction causes collapse while reducing overfitting. We propose a unified design of multi-embedding to mitigate collapse without suppressing feature interaction. Experiments on benchmark datasets demonstrate that multi-embedding consistently improves model scalability and effectively mitigates embedding collapse.


\section*{Acknowledgement}
This work was supported by the National Natural Science Foundation of China (62022050 and U2342217), the BNRist Innovation Fund (BNR2024RC01010), the Tencent Innovation Fund, and the National Engineering Research Center for Big Data Software. We also gratefully acknowledge the contributions of the following: Erheng Zhong, Xueming Qiu, Cong Li, Hui Tang, Xueyu Shi and Renjie Jiang for assistance with online deployment and experiments.

\section*{Impact Statement}
This paper presents work whose goal is to advance the deep learning research for scaling up recommendation models. There are many potential societal consequences of our work, none of which we feel must be specifically highlighted here.

\bibliography{main}
\bibliographystyle{icml2024}

\newpage
\appendix
\onecolumn

\section{Criticality of Embeddings}

For recommendation models, the embedding module occupies the largest number of parameters ($>92\%$ in our DCNv2 baseline for Criteo, and even larger for industrial models), and thus serves as the important and informative bottleneck part of the model. To further illustrate, we discuss the scaling up of the other modules of recommendation models, \textit{i.e.}, the feature interaction module $I$ and the postprocessing prediction module $F$. We experiment to increase \#cross layers and \#MLP layers in the DCNv2 baseline and show the results in Table~\ref{tab:apdx-other-module}. It is observed that increasing \#cross layers or \#MLP layers does not lead to performance improvement, hence it is reasonable and necessary to scale up the embedding size.

\begin{table}[ht]
    \centering
    \caption{Test AUC with enlarged feature interaction modules and portprocessing prediction modules. Higher indicates better.}
    \label{tab:apdx-other-module}
    \vskip 5pt
    \begin{tabular}{l|ccc}
    \toprule
    DCNv2 & 1x & 2x & 4x \\ \midrule
    standard        & 0.81339     & 0.81341     & 0.81346     \\
    + \#cross layer & 0.81325     & 0.81338     & 0.81344     \\
    + \#MLP depth   & 0.81337     & 0.81345     & 0.81342     \\ \bottomrule
    \end{tabular}
\end{table}

\section{Discussion on Embeddings in Language Models}

\begin{figure}[ht]
    \centering
    \begin{minipage}{0.9\linewidth}
        \centering
        \begin{minipage}{0.5\textwidth}
            \subfloat[DCNv2 vs. T5]{\includegraphics[width=0.9\textwidth]{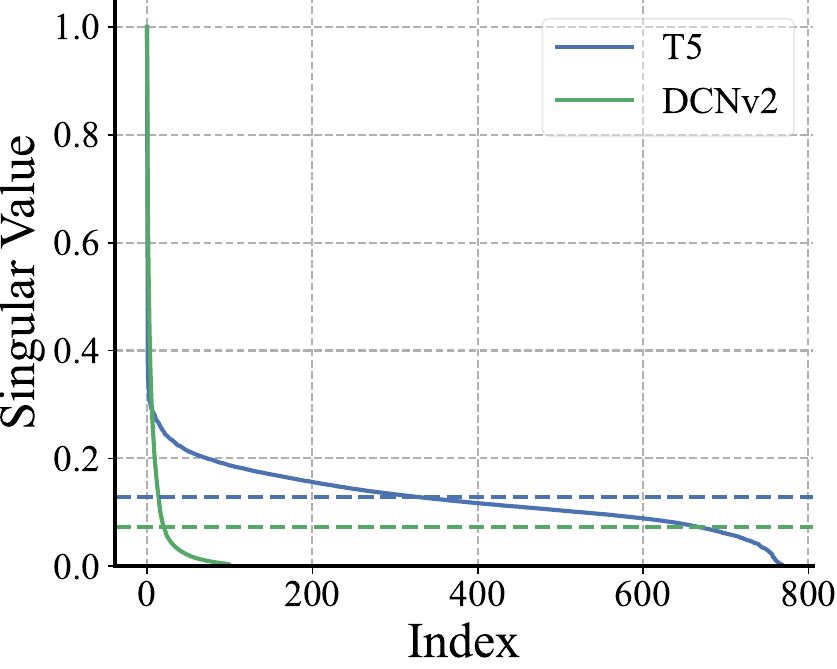}}
        \end{minipage}%
        \begin{minipage}{0.5\textwidth}
            \subfloat[DCNv2 vs. T5 (truncated)]{\includegraphics[width=0.9\textwidth]{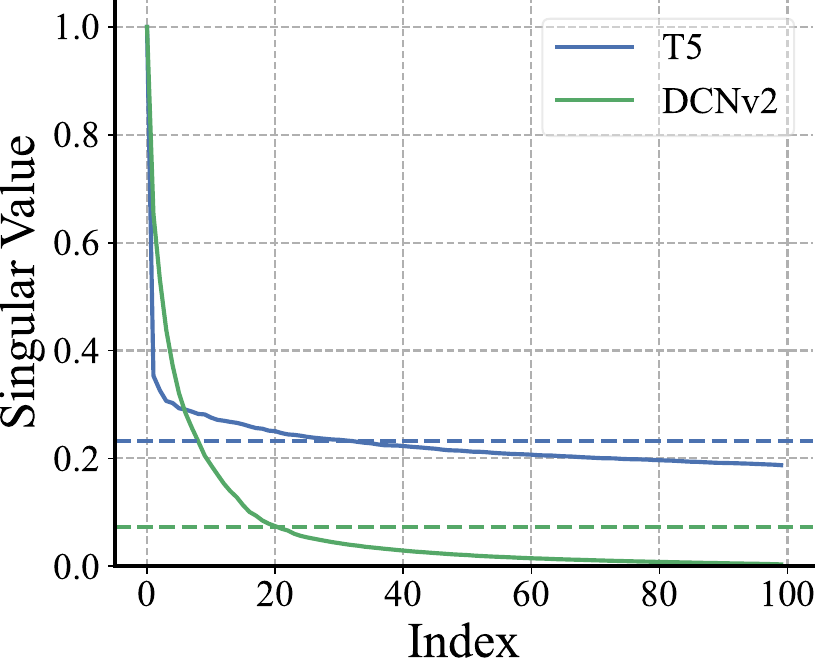}}
        \end{minipage}%
    \end{minipage}
    \caption{Comparison between T5 and DCNv2. Dash lines indicate the average singular values.}
    \label{fig:t5}
\end{figure}

To extend our analysis to other models, we examined the pretrained T5~\citep{t5} model, evaluating its (normalized) singular values for comparison. Results are shown in Figure~\ref{fig:t5}. It is observed that T5, in contrast to DCNv2, \textit{(1)} maintains higher normalized singular values and \textit{(2)} exhibits a lower proportion of insignificantly small singular values, despite its larger embedding dimensions. These observations suggest that T5 is less susceptible to the embedding collapse phenomenon, possibly because text-based models are not as affected by the interaction-collapse law in field interactions that causes embedding collapse.

\section{Details of Experiment}

\subsection{Dataset Description}

The statistics of Criteo and Avazu are shown in Table~\ref{tab:stat}. It is shown that the data amount is ample and $D_i$ can vary in a large range.

\begin{table}[ht]
    \centering
    \caption{Statistics of benchmark datasets for experiments.}
    \vskip 5pt
    \begin{tabular}{cccccc}
        \toprule
        Dataset & \#Instances & \#Fields & $\sum_i D_i$ & $\max\{D_i\}$ & $\min\{D_i\}$ \\
        \midrule
        Criteo  & 45.8M       & 39       & 1.08M      & 0.19M         & 4             \\
        Avazu   & 40.4M       & 22       & 2.02M      & 1.61M         & 5             \\
        \bottomrule
    \end{tabular}
    \label{tab:stat}
\end{table}

\subsection{Experiment Settings}
\label{apdx:exp-settings}

\paragraph{Specific multi-embedding design.} For DCNv2, DNN, IPNN and NFwFM, we add one non-linear projection after the stacked cross layers, the concatenation layer, the inner product layer and the field-weighted dot product layer, respectively. For xDeepFM, we directly average the output of the compressed interaction network, and process the ensembled DNN the same as the pure DNN model. For FinalMLP, we average the two-stream outputs respectively.

\paragraph{Hyperparameters.} For all experiments, we split the dataset into $8:1:1$ for training/validation/test with random seed 0. We use the Adam optimizer with batch size 2048, learning rate 0.001 and weight decay 1e-6. For base size, we use embedding size 50 for NFwFM considering the pooling, and 10 for all other experiments. We find the hidden size and depth of MLP does not matters the result, and for simplicity, we set hidden size to 400 and set depth to 3 (2 hidden layers and 1 output layer) for all models. We use 4 cross layers for DCNv2 and hidden size 16 for xDeepFM. All experiments use early stopping on validation AUC with patience 3. We repeat each experiment for 3 times with different random initialization. All experiments can be done with a single NVIDIA GeForce RTX 3090. 

\subsection{Experimental Results}
\label{apdx:exp-results}

\begin{figure}[ht]
    \centering
    \includegraphics[width=\textwidth]{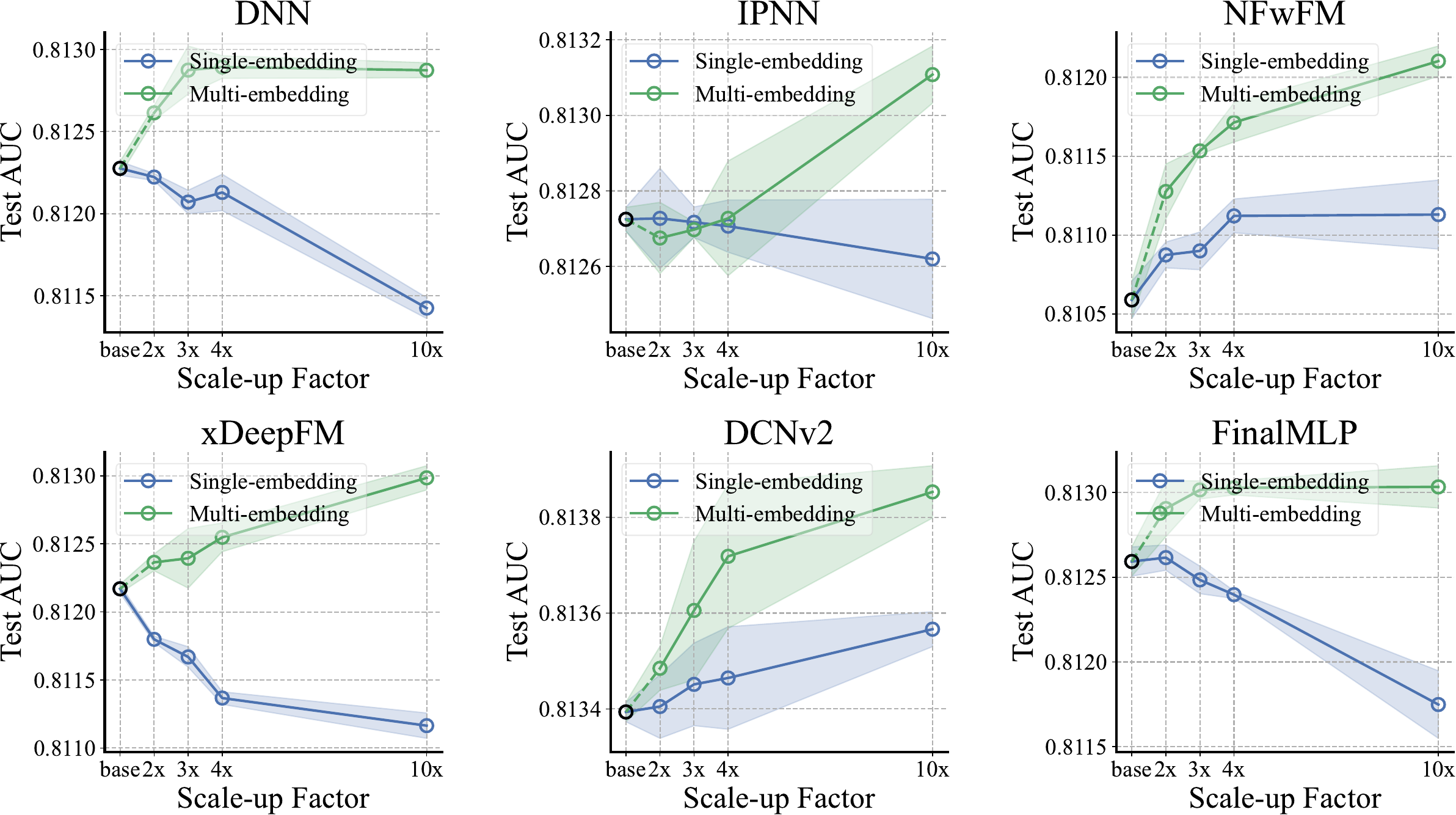}
    \caption{Visualization of scalability on Criteo dataset.}
    \label{fig:apdx-criteo}
\end{figure}

\begin{figure}[ht]
    \centering
    \includegraphics[width=\textwidth]{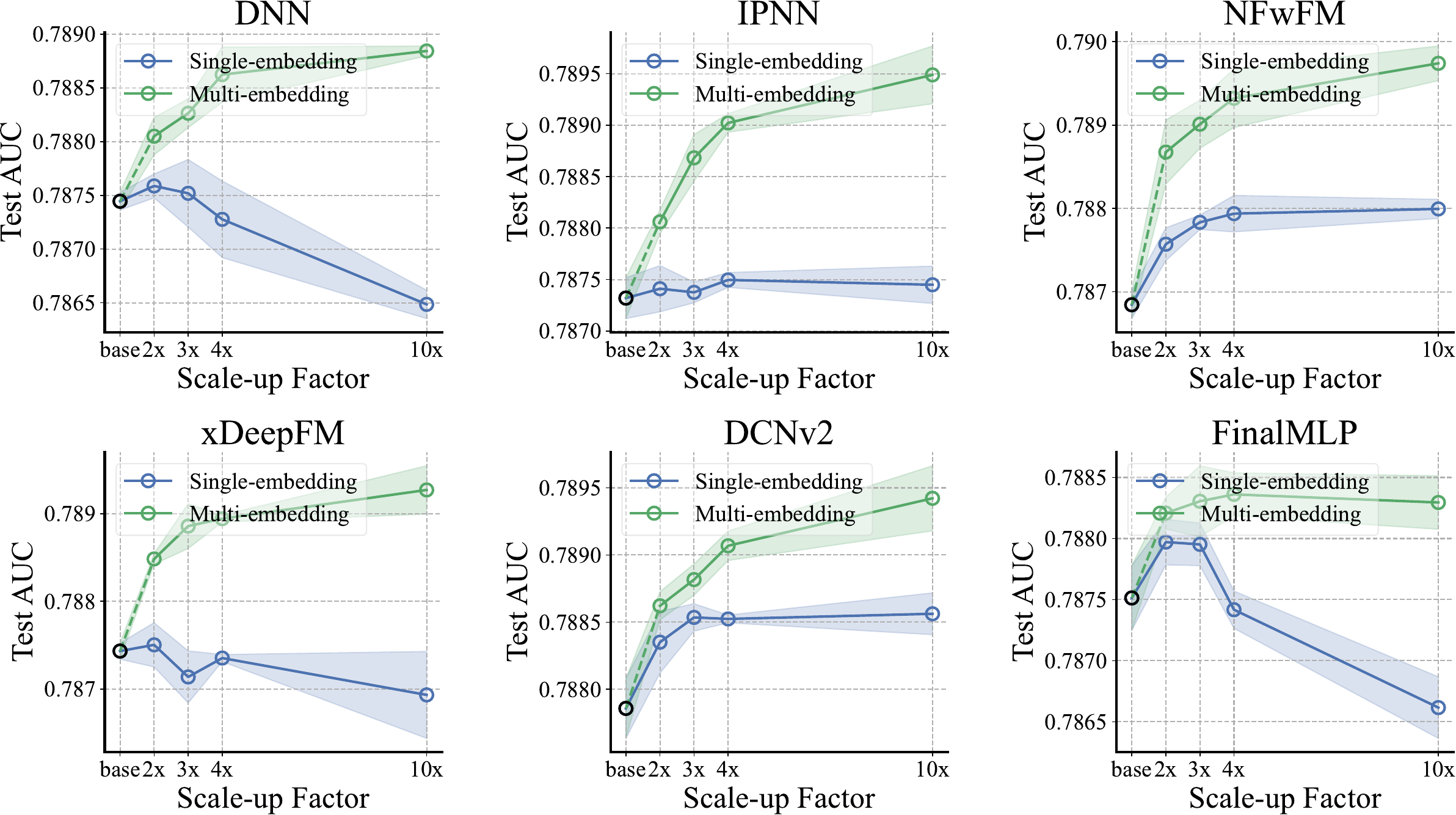}
    \caption{Visualization of scalability on Avazu dataset.}
    \label{fig:apdx-avazu}
\end{figure}

Here we present detailed experimental results with estimated standard deviation. Specifically, we show results on Criteo dataset in Table~\ref{tab:var-criteo} and Figure~\ref{fig:apdx-criteo} and Avazu dataset in Table~\ref{tab:var-avazu} and Figure~\ref{fig:apdx-avazu}.

\begin{table}[ht]
    \centering
    \caption{Results on Criteo dataset. Higher indicates better.}
    \label{tab:var-criteo}
    \vskip 5pt
            \begin{tabular}{ccccccc}
                \toprule
                \multicolumn{2}{c}{\multirow{2.5}{*}{\large Model}} & \multicolumn{5}{c}{Criteo}                                        \\ \cmidrule(lr){3-7}
                                          &                         & base                       & 2x      & 3x      & 4x      & 10x      \\ \midrule
                \multirow{2}{*}{DNN}      & SE                  & \multirow{2}{*}{0.81228$_{\pm 0.00004}$} & 0.81222$_{\pm 0.00002}$ & 0.81207$_{\pm 0.00007}$ & 0.81213$_{\pm 0.00011}$ & 0.81142$_{\pm 0.00006}$  \\
                                          & ME                &                          & 0.81261$_{\pm 0.00004}$ & 0.81288$_{\pm 0.00015}$ & 0.81289$_{\pm 0.00007}$ & 0.81287$_{\pm 0.00005}$  \\ \midrule
                \multirow{2}{*}{IPNN}     & SE                  & \multirow{2}{*}{0.81272$_{\pm 0.00003}$} & 0.81273$_{\pm 0.00013}$ & 0.81272$_{\pm 0.00004}$ & 0.81271$_{\pm 0.00007}$ & 0.81262$_{\pm 0.00016}$  \\
                                          & ME                &                          & 0.81268$_{\pm 0.00009}$ & 0.81270$_{\pm 0.00002}$ & 0.81273$_{\pm 0.00015}$ & 0.81311$_{\pm 0.00008}$  \\ \midrule
                \multirow{2}{*}{NFwFM}    & SE                  & \multirow{2}{*}{0.81059$_{\pm 0.00012}$} & 0.81087$_{\pm 0.00008}$ & 0.81090$_{\pm 0.00012}$ & 0.81112$_{\pm 0.00011}$ & 0.81113$_{\pm 0.00022}$  \\
                                          & ME                &                          & 0.81128$_{\pm 0.00017}$ & 0.81153$_{\pm 0.00002}$ & 0.81171$_{\pm 0.00012}$ & 0.81210$_{\pm 0.00010}$  \\ \midrule
                \multirow{2}{*}{xDeepFM}  & SE                  & \multirow{2}{*}{0.81217$_{\pm 0.00003}$} & 0.81180$_{\pm 0.00002}$ & 0.81167$_{\pm 0.00008}$ & 0.81137$_{\pm 0.00005}$ & 0.81116$_{\pm 0.00009}$  \\
                                          & ME                &                          & 0.81236$_{\pm 0.00006}$ & 0.81239$_{\pm 0.00022}$ & 0.81255$_{\pm 0.00011}$ & 0.81299$_{\pm 0.00009}$  \\ \midrule
                \multirow{2}{*}{DCNv2}    & SE                  & \multirow{2}{*}{0.81339$_{\pm 0.00002}$} & 0.81341$_{\pm 0.00007}$ & 0.81345$_{\pm 0.00009}$ & 0.81346$_{\pm 0.00011}$ & 0.81357$_{\pm 0.00004}$  \\
                                          & ME                &                          & 0.81348$_{\pm 0.00005}$ & 0.81361$_{\pm 0.00014}$ & 0.81382$_{\pm 0.00015}$ & 0.81385$_{\pm 0.00005}$  \\ \midrule
                \multirow{2}{*}{FinalMLP} & SE                  & \multirow{2}{*}{0.81259$_{\pm 0.00009}$} & 0.81262$_{\pm 0.00007}$ & 0.81248$_{\pm 0.00008}$ & 0.81240$_{\pm 0.00002}$ & 0.81175$_{\pm 0.00020}$  \\
                                          & ME                &                          & 0.81290$_{\pm 0.00017}$ & 0.81302$_{\pm 0.00005}$ & 0.81303$_{\pm 0.00004}$ & 0.81303$_{\pm 0.00012}$  \\ \bottomrule
            \end{tabular}
\end{table}

\begin{table}[ht]
    \centering
    \caption{Results on Avazu dataset. Higher indicates better.}
    \label{tab:var-avazu}
    \vskip 5pt
            \begin{tabular}{ccccccc}
                \toprule
                \multicolumn{2}{c}{\multirow{2.5}{*}{\large Model}} & \multicolumn{5}{c}{Avazu}                                        \\ \cmidrule(lr){3-7}
                                          &                         & base                       & 2x      & 3x      & 4x      & 10x     \\ \midrule
                \multirow{2}{*}{DNN}      & SE                  & \multirow{2}{*}{0.78744$_{\pm 0.00008}$} & 0.78759$_{\pm 0.00011}$ & 0.78752$_{\pm 0.00031}$ & 0.78728$_{\pm 0.00036}$ & 0.78648$_{\pm 0.00013}$ \\
                                          & ME                &                          & 0.78805$_{\pm 0.00017}$ & 0.78826$_{\pm 0.00013}$ & 0.78862$_{\pm 0.00026}$ & 0.78884$_{\pm 0.00005}$ \\ \midrule
                \multirow{2}{*}{IPNN}     & SE                  & \multirow{2}{*}{0.78732$_{\pm 0.00020}$} & 0.78741$_{\pm 0.00022}$ & 0.78738$_{\pm 0.00010}$ & 0.78750$_{\pm 0.00007}$ & 0.78745$_{\pm 0.00018}$ \\
                                          & ME                &                          & 0.78806$_{\pm 0.00012}$ & 0.78868$_{\pm 0.00023}$ & 0.78902$_{\pm 0.00009}$ & 0.78949$_{\pm 0.00028}$ \\ \midrule
                \multirow{2}{*}{NFwFM}    & SE                  & \multirow{2}{*}{0.78684$_{\pm 0.00017}$} & 0.78757$_{\pm 0.00020}$ & 0.78783$_{\pm 0.00009}$ & 0.78794$_{\pm 0.00022}$ & 0.78799$_{\pm 0.00011}$ \\
                                          & ME                &                          & 0.78868$_{\pm 0.00038}$ & 0.78901$_{\pm 0.00029}$ & 0.78932$_{\pm 0.00035}$ & 0.78974$_{\pm 0.00021}$   \\ \midrule
                \multirow{2}{*}{xDeepFM}  & SE                  & \multirow{2}{*}{0.78743$_{\pm 0.00009}$} & 0.78750$_{\pm 0.00025}$ & 0.78714$_{\pm 0.00030}$ & 0.78735$_{\pm 0.00004}$ & 0.78693$_{\pm 0.00050}$ \\
                                          & ME                &                          & 0.78848$_{\pm 0.00006}$ & 0.78886$_{\pm 0.00026}$ & 0.78894$_{\pm 0.00004}$ & 0.78927$_{\pm 0.00019}$ \\ \midrule
                \multirow{2}{*}{DCNv2}    & SE                  & \multirow{2}{*}{0.78786$_{\pm 0.00022}$} & 0.78835$_{\pm 0.00023}$ & 0.78854$_{\pm 0.00010}$ & 0.78852$_{\pm 0.00003}$ & 0.78856$_{\pm 0.00016}$ \\
                                          & ME                &                          & 0.78862$_{\pm 0.00011}$ & 0.78882$_{\pm 0.00012}$ & 0.78907$_{\pm 0.00011}$ & 0.78942$_{\pm 0.00024}$ \\ \midrule
                \multirow{2}{*}{FinalMLP} & SE                  & \multirow{2}{*}{0.78751$_{\pm 0.00026}$} & 0.78797$_{\pm 0.00019}$ & 0.78795$_{\pm 0.00017}$ & 0.78742$_{\pm 0.00015}$ & 0.78662$_{\pm 0.00025}$ \\
                                          & ME                &                          & 0.78821$_{\pm 0.00013}$ & 0.78831$_{\pm 0.00029}$ & 0.78836$_{\pm 0.00018}$ & 0.78830$_{\pm 0.00022}$ \\ \bottomrule
            \end{tabular}
\end{table}

\section{More Baseline Methods}

We also conduct experiments on AutoInt~\citep{autoint} and compare the performance under single-embedding and multi-embedding. Due to the limited computational resource, we only scale up the model to 4x on Criteo dataset. Results are shown in Table~\ref{tab:autoint}. It is observed that single-embedding suffers from non-scalability while our multi-embedding consistently enhance performance along with the model size, achieving 6e-4 AUC improvement by simply scaling up.

\begin{table}[ht]
\centering
\caption{Results on Criteo dataset. Higher indicates better.}
\begin{tabular}{c|cccc}
\toprule
AutoInt & base & 2x & 3x & 4x \\ \midrule
SE         & \multirow{2}{*}{0.81231}       & 0.81236     & 0.81216     & 0.81193     \\
ME         &               & 0.81264     & 0.81285     & 0.81291     \\
\bottomrule
\end{tabular}
\label{tab:autoint}
\end{table}

\section{Non-Linearity for Multi-Embedding}
\label{apdx:non-linearity}
We have mentioned that the embedding-set-specific feature interaction of multi-embedding should contain non-linearity, otherwise the model will degrade to a single-embedding model. For simplicity, we consider a stronger version of multi-embedding, where the combined features from different embedding sets are concatenated instead of averaged. To further illustrate, consider linear feature interaction modules $I^{(m)}:\left(\mathbb{R}^K\right)^N\to\mathbb{R}^{h}$, then we can define a linear feature interaction module $I_{\text{all}}:\left(\mathbb{R}^{MK}\right)^N\to\mathbb{R}^{Mh}$. For convenience, we denote $[f(i)]_{i=1}^n$ as $[f(1),f(2),...,f(n)]$, and $\ve_i=[\ve_i^{m}]_{m=1}^M$. The form of $I_{\text{all}}$ can be formulated by
\[
    I_{\text{all}}\left(\ve_1,\ve_2,...,\ve_N\right)=\left[I^{(m)}(\ve_1^{(m)},...,\ve_N^{(m)})\right]_{m=1}^M.
\]
This shows a multi-embedding model is equivalent to a model by concatenating all embedding sets. We will further show that the deduced model with $I_{\text{all}}$ is homogeneous to a single-embedding model with size $MK$, \textit{i.e.}, multi-embedding is similar to single-embedding with linear feature interaction modules. Denote the feature interaction module of single-embedding as $I$. Despite $I_{\text{all}}$ could have different forms from $I$, we further give three examples to show the homogeneity of $I_{\text{all}}$ and $I$.

\paragraph{DNN.} Ignoring the followed MLP, DNN incorporate a non-parametric interaction module by concatenating all fields together. Formally, we have
\begin{align*}
    I(\ve_1,...,\ve_N)&=\left[[\ve_i^{(m)}]_{m=1}^M\right]_{i=1}^N, \\
    I_{\text{all}}(\ve_1,...,\ve_N)&=\left[[\ve_i^{(m)}]_{i=1}^N\right]_{m=1}^M.
\end{align*}
In other words, $I$ and $I_{\text{all}}$ only differ in a permutation, thus multi-embedding and single-embedding are equivalent.

\paragraph{Projected DNN.} If we add a linear projection after DNN, then we can split the projection for fields and embedding sets, and derive
\begin{align*}
    I(\ve_1,...,\ve_N)&=\sum_{i=1}^N\sum_{m=1}^M\mW_{i,m}\ve_i^{(m)}, \\
    I_{\text{all}}(\ve_1,...,\ve_N)&=\left[\sum_{i=1}^N\mW_{i,m}\ve_i^{(m)}\right]_{m=1}^M.
\end{align*}
In other words, $I$ and $I_{\text{all}}$ only differ in a summation. Actually if we average the combined features for $I_{\text{all}}$ rather than concatenate to restore our proposed version of multi-embedding, then multi-embedding and single-embedding are equivalent by the scalar $1/M$.

\paragraph{DCNv2.} DCNv2 incorporates the following feature interaction by
\[
    I(\ve_1,...,\ve_N)=\left[\ve_i\odot\sum_{j=1}^N\mW_{j\to i}\ve_j\right]_{i=1}^N,
\]
thus by splitting $\mW_{i\to j}$ we have
\begin{align*}
    I(\ve_1,...,\ve_N)&=\left[[\ve_i^{(m)}\odot\sum_{j=1}^N\sum_{m'=1}^M\mW^{(m,m')}_{j\to i}\ve^{(m')}_j]_{m=1}^M\right]_{i=1}^N \\
    I_{\text{all}}(\ve_1,...,\ve_N)&=\left[[\ve^{(m)}_i\odot\sum_{j=1}^N\mW^{(m)}_{j\to i}\ve^{(m)}_j]_{i=1}^N\right]_{m=1}^M.
\end{align*}
By simply letting $\mW^{(m,m)}=\mW^{(m)}$ and $\mW^{(m,m')}=\mO$ for $m\neq m'$, we convert a multi-embedding model into a single-embedding model under permutation. Therefore, multi-embedding is a special case of single-embedding for DCNv2.

\paragraph{Summary.} In summary, a linear feature interaction module will cause homogenity between single-embedding and multi-embedding. Hence it is necessary to use or introduce non-linearity in feature interaction module.

\section{Detailed Explanation of Embedding Diversity}
\label{apdx:diversity_theory}

In Section~\ref{sec:how_works}, we propose to use principal angle to measure embedding set diversity. Here we introduce the motivation and an example. Note that
\begin{align*}
\mathrm{rank}\left(\left[\mE^{(m)},\mE^{(m')}\right]\right)
&=\mathrm{rank}\left(\left[\mU^{(m)}\mSigma^{(m)}\big(\mV^{(m)}\big)^\top,\mU^{(m')}\mSigma^{(m')}\big(\mV^{(m')}\big)^\top\right]\right) \\
&=\mathrm{rank}\left(\left[\mU^{(m)}\mSigma^{(m)}\big(\mV^{(m)}\big)^\top,\mU^{(m')}\mSigma^{(m')}\big(\mV^{(m')}\big)^\top\right]\begin{bmatrix}
    \mV^{(m)} & \mO \\
    \mO & \mV^{(m')}
\end{bmatrix}\right) \\
&=\mathrm{rank}\left(\left[\mU^{(m)}\mSigma^{(m)},\mU^{(m')}\mSigma^{(m')}\right]\right) \\
&=\mathrm{rank}\left(\mU^{(m)}\mSigma^{(m)}\right)+\mathrm{rank}\left(\mU^{(m')}\mSigma^{(m')}\right)-\mathrm{rank}\left(\left(\mU^{(m)}\mSigma^{(m)}\right)^\top \mU^{(m')}\mSigma^{(m')}\right) \\
&=\mathrm{rank}\left(\mE^{(m)}\right)+\mathrm{rank}\left(\mE^{(m')}\right)-\mathrm{rank}\left(\big(\mU^{(m)}\big)^\top \mU^{(m')}\right),
\end{align*}
where the second last line are derived from the orthonormality of $\mU$. Note that
\[
\mathrm{rank}\left(\big(\mU^{(m)}\big)^\top \mU^{(m')}\right)=\|\cos\left(\bm{\phi}^{m\leftrightarrow m'}\right)\|_0,
\]
and we therefore generalize it to $\frac1K\|\cos\left(\bm{\phi}^{m\leftrightarrow m'}\right)\|_1$ to measure similarity, and use $1-\mathrm{similarity}$ as diversity.

Considering the following example of diversity. An embedding with size of $2$ are learned as
\[
    \mE=\begin{pmatrix}
        1&0 \\
        1&0 \\
        0&1 \\
        0&1
    \end{pmatrix}
\]
with $\mathrm{rank}(\mE)=\mathrm{IA}(\mE)=2$. If enlarged to size of $4$, due to interaction-collapse theory, it is likely to be learned as 
\[
    \mE^{(1)}=\begin{pmatrix}
        1&0 \\
        1&0 \\
        0&1 \\
        0&1
    \end{pmatrix},\quad
    \mE^{(2)}=\begin{pmatrix}
        1&0 \\
        1&0 \\
        0&1 \\
        0&1
    \end{pmatrix},\quad
    \mE^{\text{single}}=\begin{pmatrix}
        1&0&1&0 \\
        1&0&1&0 \\
        0&1&0&1 \\
        0&1&0&1
    \end{pmatrix},
\]
with $\cos\left(\bm{\phi}^{1\leftrightarrow 2}\right)=(1,1)$, $\mathrm{rank}(\mE^{\text{single}})=\mathrm{IA}(\mE^{\text{single}})=2$, \textit{i.e.}, the enlarged size does not increase information abundance. Here 

When using multi-embedding, the embedding sets are possibly learned to be of large diversity, and the overall embedding are learned as
\[
    \mE^{(1)}=\begin{pmatrix}
        1&0 \\
        1&0 \\
        0&1 \\
        0&1
    \end{pmatrix},\quad
    \mE^{(2)}=\begin{pmatrix}
        1&0 \\
        0&1 \\
        1&0 \\
        0&1
    \end{pmatrix},\quad
    \mE^{\text{multi}}=\begin{pmatrix}
        1&0&1&0 \\
        1&0&0&1 \\
        0&1&1&0 \\
        0&1&0&1
    \end{pmatrix},
\]
with $\cos\left(\bm{\phi}^{1\leftrightarrow 2}\right)=(1,0)$, $\mathrm{rank}(\mE^{\text{multi}})=3$ and $\mathrm{IA}(\mE^{\text{multi}})=1+\sqrt2$, indicating the effectiveness of multi-embedding.

\section{Details of Toy Expermient}
\label{apdx:toy}

In this section, we present the detailed settings of the toy experiment. We consider a scenario with $N=3$ fields and $D_1=D_2=100$. For each $(x_1,x_2)\in \mathcal{X}_1\times\mathcal{X}_2$, we randomly assign $x_3\sim\mathcal{U}[\mathcal{X}_3]$, $y\sim\mathcal{U}\{0,1\}$ and let $(\vx,y)$ to be one piece of data, thus for different values of $D_3$, there are always $100^2$ pieces of data, and they follow the same distribution when reduced on $\mathcal{X}_1\times\mathcal{X}_2$. We set $D_3=3$ and $D_3=100$ to simulate the case with low-information-abundance and high-information-abundance, respectively. We randomly initialize all embedding matrices with normal distribution $\mathcal{N}(0,1)$, fix $\mE_2,\mE_3$ and only optimize $\mE_1$ during training. We use full-batch SGD with the learning rate of 1. We train the model for 5,000 iterations in total.

\section{Empirical analysis on FFM in Evidence I}
\label{apdx:ffm}
\begin{figure}[ht]
    \centering
    \begin{minipage}{0.3333\textwidth}
        \centering
        \subfloat[$\mathrm{IA}(\mE_i^{\to j})$.\label{subfig:evd1a}]{\includegraphics[width=0.85\textwidth]{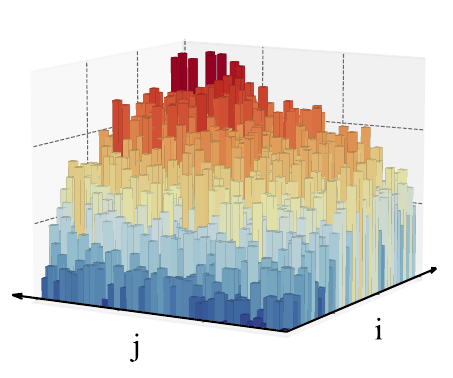}}
    \end{minipage}%
    \begin{minipage}{0.3333\textwidth}
        \centering
        \subfloat[$\sum\limits_{j=1}^N\mathrm{IA}(\mE_i^{\to j})$.\label{subfig:evd1b}]{\includegraphics[width=0.85\textwidth]{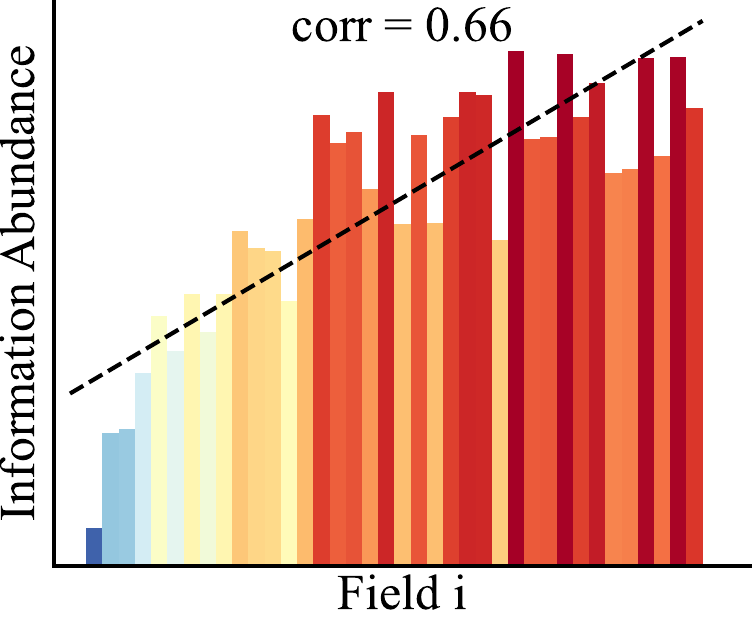}}
    \end{minipage}%
    \begin{minipage}{0.3333\textwidth}
        \centering
        \subfloat[$\sum\limits_{i=1}^N\mathrm{IA}(\mE_i^{\to j})$.\label{subfig:evd1c}]{\includegraphics[width=0.85\textwidth]{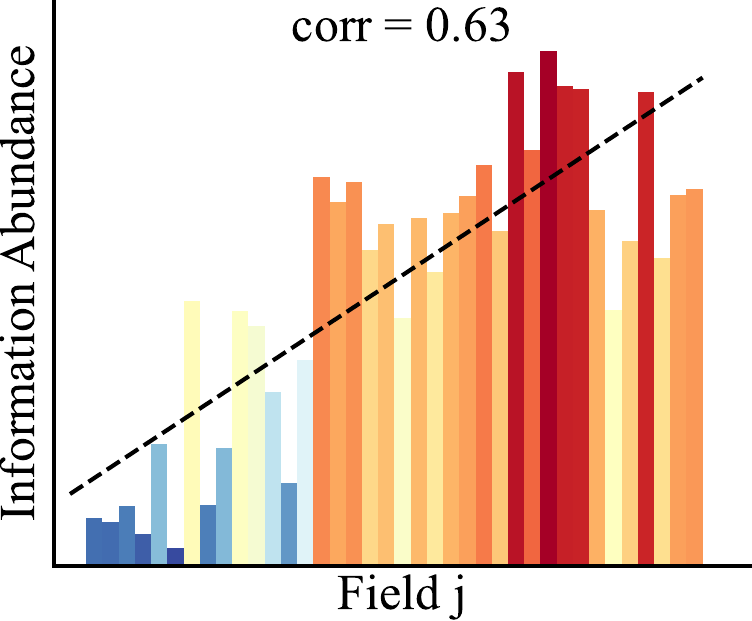}}
    \end{minipage}
    \caption{Information abundance of sub-embedding matrices for FFM, with field indices sorted by information abundance of corresponding raw embedding matrices. Higher or warmer indicates larger. Similarly, $\mathrm{IA}(\mE_i^{\to j})$ are co-influenced by both $\mathrm{IA}(\mE_i)$ and $\mathrm{IA}(\mE_j)$.}
    \label{fig:evd12}
\end{figure}

Field-aware factorization machines (FFM)~\citep{ffm} split an embedding matrix of field $i$ into multiple sub-embeddings with
\[
\mE_i=\left[\mE_i^{\to 1},\mE_i^{\to 2},...,\mE_i^{\to(i-1)},\mE_i^{\to(i+1)},...,\mE_i^{\to N}\right],
\]
where sub-embedding $\mE_i^{\to j}\in\mathbb{R}^{D_i\times K/(N-1)}$ is only used when interacting field $i$ with field $j$ for $j\neq i$. We perform the same experiments as Evidence I, and similarly find that $\mathrm{IA}(\mE_i^{\to j})$ are co-influenced by both $\mathrm{IA}(\mE_i)$ and $\mathrm{IA}(\mE_j)$, as shown in Figure~\ref{fig:evd12}. This is amazing in the sense that even using independent embeddings to represent the same field features, these embeddings get different information abundance after learning.

\section{Extension of Information Abundance}

\begin{figure}[ht]
    \centering
    \includegraphics[width=0.3\textwidth]{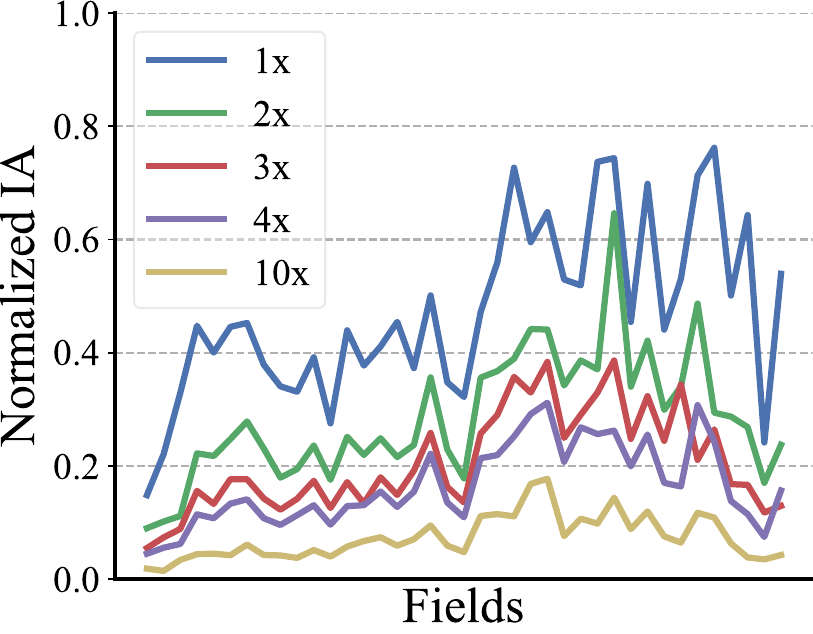}
    \caption{Normalized information abundance $\frac{\mathrm{IA}(\mE)}{K}$ for different embedding sizes on DCNv2.}
    \label{fig:apdx-normalized-ia}
\end{figure}

Our proposed information abundance is a fair metric when two embedding matrices have the same embedding size. To apply the definition between different embedding sizes, some possible extensions include
$\frac{\mathrm{IA}(\mE)}{K}$ and $\frac{\mathrm{IA}(\mE)}{\mathbb{E}[\mathrm{IA}(\texttt{randn\_like}(\mE))]}$, where $K$ stands for the embedding size and $\texttt{randn\_like}(\mE)$ refers to a random matrix underlying the normal distribution with the same shape as $\mE$. We compare the former one with different embedding sizes in Figure~\ref{fig:apdx-normalized-ia}, and it is shown that the degree of collapse increases with respect to the embedding size, which is consistent with the observation in Figure~\ref{subfig:intro-b}

\section{Detailed Explanation of Regularized DCNv2}

\label{apdx:rdcnv2}

\begin{figure}[ht]
    \centering
    \begin{minipage}{0.9\textwidth}
        \centering
        \begin{minipage}{0.5\textwidth}
            \centering
            \subfloat[Standard DCNv2]{\includegraphics[width=0.8\textwidth]{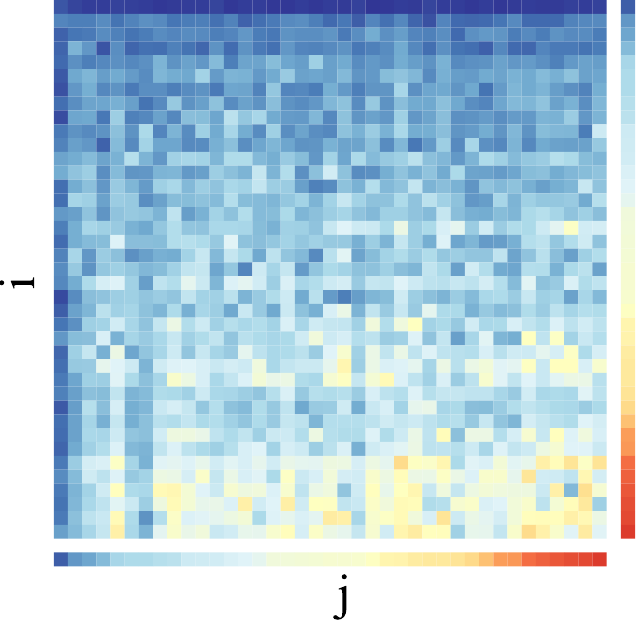}}
        \end{minipage}%
        \begin{minipage}{0.5\textwidth}
            \centering
            \subfloat[Regularized DCNv2]{\includegraphics[width=0.8\textwidth]{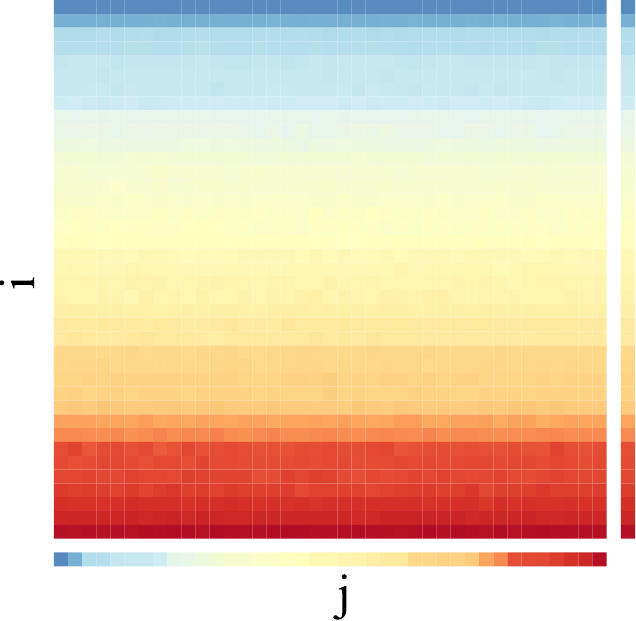}}
        \end{minipage}
    \end{minipage}
    \caption{Visualization of information abundance of embeddings and sub-embeddings for standard and regularized DCNv2, respectively. The rightmost and downmost bars correspond to $\mathrm{IA}(\mE_i)$ or $\mathrm{IA}(\mE_j)$. Compared with the standard DCNv2, the regularized can preserve singular values, and resulting in less-collapsed sub-embeddings and finally larger information abundance as in Figure~\ref{subfig:evd3}.}
    \label{fig:apdx-heatmap}
\end{figure}

Regarding Evidence II, we proposed regularization of the weight matrix $\mW_i^{\to j}$ to mitigate the collapse caused by the projection $\mW_i^{\to j}$ in sub-embeddings. By regularizing $\mW_i^{\to j}$ to be a unitary matrix (or the multiplication of unitary matrices), we ensure the preservation of all singular values of the sub-embedding. Consequently, the information abundance of sub-embeddings in regularized DCNv2 is larger than standard DCNv2. We plot the heatmap of information abundance of embeddings and sub-embeddings Figure~\ref{fig:apdx-heatmap}. This clearly demonstrates that regularized DCNv2 exhibits a higher information abundance. Based on our Finding 1, regularized DCNv2 mitigates the problem of embedding collapse by increasing the information abundance of the sub-embeddings that directly interact with the embeddings.

\section{How ME Performs When Feature Interaction Is Suppressed?}

In this section we analysis of ME under models with feature interaction suppressed, as discussed Section~\ref{subsec:generalizability}, where SE suffers from overfiting.

\begin{figure}[ht]
    \centering
    \subfloat[Standard vs Regularized\label{fig:evd3apdx}]{\includegraphics[width=0.45\textwidth]{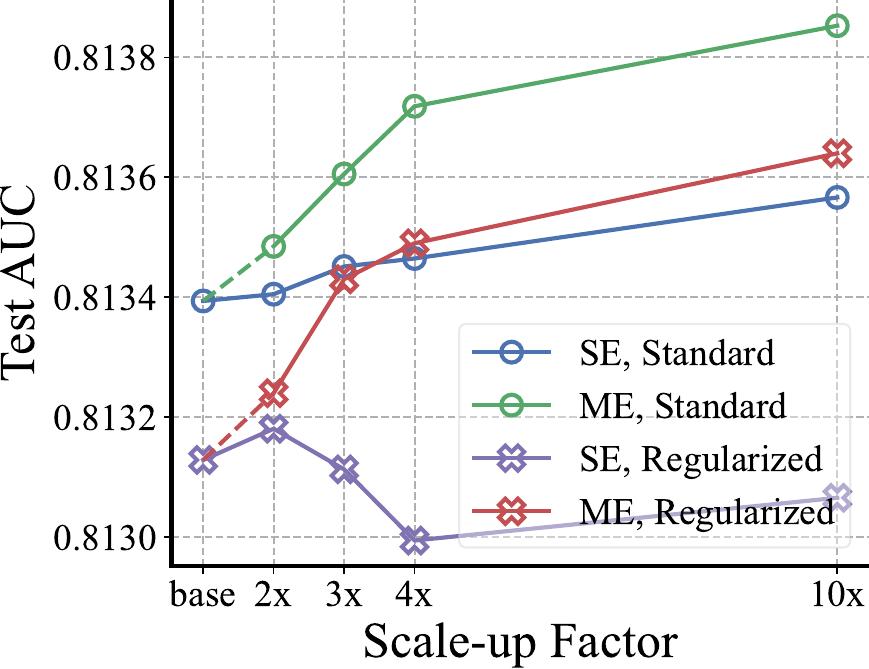}}
    \subfloat[DCNv2 vs DNN\label{fig:evd4apdx}]{\includegraphics[width=0.45\textwidth]{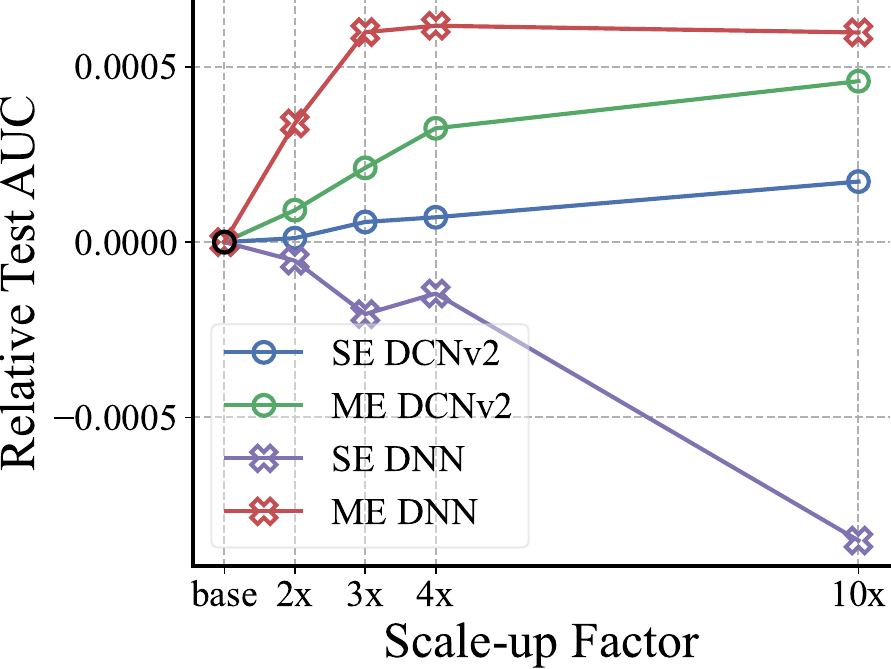}}
    \caption{Test AUC w.r.t. model size.}
\end{figure}

\paragraph{Evidence III for ME.} We add regularization
\[
    \ell_{reg}=\sum_{m=1}^M\sum_{i=1}^N\sum_{j=1}^N\left\|(\mW^{(m)}_{i\to j})^\top \mW^{(m)}_{i\to j}-\lambda^{(m)}_{ij} \mI\right\|_{\mathrm{F}}^2
\]
for ME DCNv2 and conduct experiments with different embedding sizes. Results are shown in Figure~\ref{fig:evd3apdx}. Even though the performance is worse than that without regularization, compared with SE, ME still consistently improves the performance with the growth of model size.

\paragraph{Evidence IV for ME.} We compare the performance of SE/ME on DNN/DCNv2, as shown in Figure~\ref{fig:evd4apdx}. Compared with SE, ME DNN improves the performance with the growth of model size.

\paragraph{Summary: ME offers scalability even though feature interaction is suppressed.} For models with feature interaction suppressed such as regularized DCNv2 and DNN, the performance of SE might dropped when enlarging models, since feature interaction provides domain knowledge and large models might suffer from overfitting. Experiments show that these models with ME can properly scale up. Such results are plausible since ME improves scalability by capturing diverse patterns instead of learning with a single interaction pattern.





\end{document}